%% file: ijcai26.tex
\documentclass{article}
\usepackage{ijcai26}

\usepackage{multirow}
\usepackage{times}
\usepackage{soul}
\usepackage{url}
\usepackage[hidelinks]{hyperref}
\usepackage[utf8]{inputenc}
\usepackage[small]{caption}
\usepackage{graphicx}
\usepackage{amsmath}
\usepackage{amsthm}
\usepackage{amssymb} 
\usepackage[inline]{enumitem}
\usepackage{booktabs}
\usepackage{algorithm}
\usepackage{algorithmic}
\usepackage[switch]{lineno}
\usepackage[hidelinks]{hyperref}
\usepackage[nameinlink,noabbrev]{cleveref}
\usepackage[table,x11names,dvipsnames,table]{xcolor}
\usepackage[round]{natbib}
\usepackage{arydshln}

\newtheorem{theorem}{Theorem}
\newtheorem{lemma}[theorem]{Lemma}

\newcommand{\DashedMidrule}{%
    \noalign{\vskip 0.5ex}  
    \hdashline
    \noalign{\vskip 0.5ex}  
}

\title{FLASH: Flexible Learning of Adaptive Sampling
from History in Temporal Graph Neural Networks}

\author{
    Or Feldman\textsuperscript{1},
    Krishna Sri Ipsit Mantri\textsuperscript{2},
    Carola-Bibiane Schönlieb\textsuperscript{3}\\
    Chaim Baskin\textsuperscript{1}\thanks{Equal supervision.},
    Moshe Eliasof\textsuperscript{1,3}\footnotemark[1]\\
    \affiliations
    \textsuperscript{1}Ben-Gurion University of the Negev, Israel\\  \textsuperscript{2}Purdue University, USA\\   \textsuperscript{3}University of Cambridge, UK 
    \emails
    orfel@post.bgu.ac.il, kmantri@uni-bonn.de, \ cbs31@cam.ac.uk, chaimbaskin@bgu.ac.il eliasof@bgu.ac.il
}

\newcommand{\ourmethod}{\textsc{FLASH}}
\newcommand{\methodname}{\textsc{FLASH}}
\usepackage[round]{natbib}

\usepackage{pifont}
\newcommand{\cmark}{\ding{51}}%
\newcommand{\xmark}{\ding{55}}%

\input{math_commands}

\begin{document}

\maketitle

\input{ijcai26/sections/abstract}
\input{ijcai26/sections/Introduction}

\input{ijcai26/sections/RelatedWorks}
\input{ijcai26/sections/Background}

\input{ijcai26/sections/Method}

\input{ijcai26/sections/Experiments}

\input{ijcai26/sections/Conclusion}

\clearpage

\bibliographystyle{ijcai26}
\bibliography{ijcai26}

\appendix

\include{ijcai26/appendix/AppendixA}
\include{ijcai26/appendix/AppendixB}
\include{ijcai26/appendix/AppendixC}

\include{ijcai26/appendix/appendixD}
\include{ijcai26/appendix/appendixE}
\include{ijcai26/appendix/appendixF}
\include{ijcai26/appendix/AppendixG}




\end{document}

%% file: math_commands.tex
\usepackage{amsmath,amsfonts,bm}

\def\eqref#1{equation~\ref{#1}}

\def\1{\bm{1}}

\def\vtheta{{\bm{\theta}}}

\def\vh{{\bm{h}}}

\def\vz{{\bm{z}}}

\DeclareMathAlphabet{\mathsfit}{\encodingdefault}{\sfdefault}{m}{sl}
\SetMathAlphabet{\mathsfit}{bold}{\encodingdefault}{\sfdefault}{bx}{n}

\def\gE{{\mathcal{E}}}
\def\gF{{\mathcal{F}}}
\def\gG{{\mathcal{G}}}
\def\gH{{\mathcal{H}}}

\def\gM{{\mathcal{M}}}

\def\gV{{\mathcal{V}}}

\def\sR{{\mathbb{R}}}



%% file: ijcai26/sections/abstract.tex
\begin{abstract}
Aggregating temporal signals from historic interactions is a key step in future link prediction on dynamic graphs. However, incorporating long histories is resource-intensive. Hence, temporal graph neural networks (TGNNs) often rely on historical neighbors sampling heuristics such as uniform sampling or recent neighbors selection. These heuristics are static and fail to adapt to the underlying graph structure. We introduce \ourmethod{}, a learnable and graph-adaptive neighborhood selection mechanism that generalizes existing heuristics. \ourmethod{} integrates seamlessly into TGNNs and is trained end-to-end using a self-supervised ranking loss. We provide theoretical evidence that commonly used heuristics hinder TGNNs performance, motivating our design. Extensive experiments across multiple benchmarks demonstrate consistent and significant performance improvements for TGNNs equipped with \ourmethod{}.
\end{abstract}

%% file: ijcai26/sections/Introduction.tex
\section{Introduction}
\label{sec:introduction}
Dynamic graphs provide a natural framework for modeling real-world systems where entities and their interactions evolve over time. They underpin a wide range of applications, including social and communication networks~\cite{shetty2004enron,kumar2019predicting}, user-item recommendation systems~\cite{kumar2019predicting}, and financial or knowledge-intensive platforms~\cite{ni2019justifying,shamsi2022chartalist}. Predicting future interactions in these settings has emerged as a central learning task, leading to the development of Temporal Graph Neural Networks (TGNNs) -- models specifically designed to learn from sequences of timestamped events represented by dynamic graphs.

\begin{figure}[t]
\centering
\includegraphics[width=0.75\linewidth]{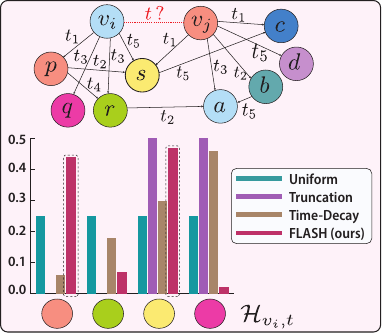}

\caption{%
Illustration of different neighborhood selection strategies for predicting a link between $v_i$ and $v_j$. 
Circles represent nodes and their colors indicate each node’s feature. 
The bar chart on the right shows how each strategy scores the neighbors. 
Static heuristics (truncation or uniform sampling) either discard them or fail to prioritize important neighbors, while \ourmethod{} assigns higher scores to these key neighbors. 
}

\label{fig:teaser}
\end{figure}
TGNNs process dynamic graphs by encoding temporal interaction patterns into node representations, enabling them to predict future links. A common challenge in these models is how to efficiently aggregate information from a node's history of interactions, which can grow unbounded over time. Processing complete histories quickly becomes computationally prohibitive, especially in high-frequency interaction settings. To address this, existing models such as TGN~\cite{rossi2020temporal}, TGAT~\cite{xu2020inductive}, DyGFormer~\cite{yu2023towards}, GraphMixer~\cite{cong2023we}, FreeDyG~\cite{tian2023freedyg}, and others~\cite{zhou2022tgl,zou2024repeat,gravina2024long,xu2024timesgn} adopt memory-efficient heuristics. These typically include strategies like uniform sampling, time-decay weighting, or truncating to the $k$ most recent interactions. While effective at reducing computational overhead, these approaches are static and fail to adapt to the local graph structure or task-specific temporal signals. As shown in \Cref{fig:teaser}, such heuristics apply uniform or truncated selection schemes that overlook potentially informative neighbors, whereas adaptive strategies can prioritize structurally meaningful interactions.

Static heuristics like uniform sampling~\cite{rossi2020temporal}, truncation~\cite{cong2023we} or hybrid approaches~\cite{luoscalable} are appealing due to their simplicity, but they treat all interactions as equally informative or rely solely on recency. This ignores the fact that some neighbors may be more relevant than others due to their position in the graph or their interaction patterns. Moreover, the optimal sampled neighborhood may vary across time, nodes, and tasks, making fixed strategies fundamentally limited. These shortcomings are further amplified in heterogeneous or rapidly evolving graphs, where structural context can shift dramatically over time. This motivates the need for a learnable, structure-aware neighborhood selection mechanism that can adaptively prioritize informative past interactions.

To address these limitations, we propose \ourmethod{} -- \textbf{\underline{F}lexible \underline{L}earning of \underline{A}daptive \underline{S}election from \underline{H}istory} \textbf{for Temporal Graph Neural Networks}, a learnable and graph-adaptive neighborhood selection mechanism for TGNNs. \ourmethod{} replaces static heuristics with a data-driven approach that learns to prioritize historically informative neighbors based on their structural and temporal context. Crucially, because the true importance of neighbors is not known a priori, our method is trained using a self-supervised ranking objective that encourages selecting neighbors most predictive of future interactions. \ourmethod{} is lightweight, general-purpose, and integrates seamlessly into a wide range of existing TGNN architectures, including TGNNs with non-differentiable feature extractors~\cite{yu2023towards,tian2023freedyg}.  This allows it to improve predictive performance without requiring architectural changes. Our key contributions are as follows:
\begin{itemize}
\item We propose \ourmethod{}, a novel graph-adaptive, learnable neighborhood selection mechanism that seamlessly integrates with any TGNN.
 \item We design a self-supervised training objective based on ranking loss, enabling our method to learn informative neighbor selection without access to ground-truth labels.
\item We provide a theoretical analysis showing that \ourmethod{} is provably more expressive than the common heuristics of recent neighbors selection and uniform sampling.  
\item We conduct extensive experiments across multiple dynamic graph benchmarks, demonstrating consistent performance gains across diverse TGNN backbones compared to common neighbor sampling baselines.
\end{itemize}

%% file: ijcai26/sections/RelatedWorks.tex
\section{Related Work}
\label{sec:relatedworks}

\paragraph{Neighborhood selection in static graphs} Existing sampling techniques for large static graphs often rely on substructure sampling (e.g., nodes or edges), as employed by GraphSAGE ~\cite{hamilton2017inductive} and FastGCN~\cite{chen2018fastgcn}, or utilize random walks, as in PinSage~\cite{ying2018graph}. Other methods, such as GraphSAINT~\cite{zeng2019graphsaint} and Cluster-GCN~\cite{chiang2019cluster}, are specifically designed to facilitate efficient training on large graphs. However, these approaches typically do not address inference or the temporal nature of dynamic graphs.
In many TGNNs, uniform sampling can be viewed as dynamic extension of GraphSAGE-like methods, in which each neighbor in the historical neighborhood is sampled with equal probability. On the other hand, truncation sampling retains only the most recent neighbors, and its stochastic variant of time-weighting, reduces the sampling probability of older neighbors over time, effectively treating recency as a measure of importance. This parallels importance sampling in static graphs (e.g., FastGCN), where recent interactions are explicitly prioritized. However, our experiments show that recency alone is insufficient to capture the complexities of temporal interactions. Instead, incorporating node-specific contextual information at each interaction point, as done by \ourmethod{}, proves crucial for achieving robust and accurate performance in dynamic graph settings.

\paragraph{Learning from Large Historical Neighborhoods} 
Learning from large historical neighborhoods in dynamic graphs poses significant challenges in computational cost and capturing long-term dependencies. To address the latter, the patching technique introduced in~\cite{yu2023towards}, adopted in other recent studies~\cite{tian2023freedyg,dingdygmamba}. The method splits the historical neighborhood into chronological patches, each linearly projected into a single representative vector. Since historical neighbors must be encoded (e.g., via co-occurrence encoding~\cite{yu2023towards}), neighborhood selection precedes it to reduce computational overhead. Thus, as proposed by~\cite{yu2023towards} for \textsc{DyGFormer}, patching is a complementary strategy for processing large historical neighborhoods,  in future link prediction tasks.

\paragraph{Weighing and Selection Mechanisms.}
Some approaches, such as TGAT~\cite{xu2020inductive}, employ attention mechanisms to weight neighbors, but only after the neighborhood has been sampled. Other works use reinforcement learning~\cite{wang2022reinforcementlearningenhancedweighted} to learn selection strategies, but they require task-specific reward design and extensive training. Methods based on hard negative mining or curriculum sampling~\cite{younesian2024grapeslearningsamplegraphs} share our motivation of identifying informative samples, but are not tailored for self-supervised learning and are not readily applicable to TGNNs. In contrast, \ourmethod{} learns to sample informative neighbors in a self-supervised, fully differentiable manner.

%% file: ijcai26/sections/Background.tex
\section{Background}
\label{sec:background}
Continuous Time Dynamic Graph (CTDG) is represented as a sequence of time-stamped events $\gG = \{(x_1,t_1), (x_2,t_2) \cdots\}$ where $t_1 \leq t_2 \leq \cdots$. Each event $(x_i, t_i)$ represents an action on the graph that occurred at time $t_i$. Each such action can be either node addition, node removal, edge addition, or edge removal. We denote $\gG_t = (\gV_t, \gE_t)$ the snapshot of CTDG at time $t$, which is the graph received by applying all the events in $\gG$ that occurred until time $t$, where $\gG_0 = (\varnothing,\varnothing)$. We denote $F_{\gV} : \gV \times \sR^+ \to \sR^{d_{\gV}}$ and $F_{\gE} : \gE \times \sR^+ \to \sR^{d_{\gE}}$  as the functions that map a node or an edge to their features, at a specific point in time.
\begin{figure*}[t]
    \centering\includegraphics[width=\linewidth]{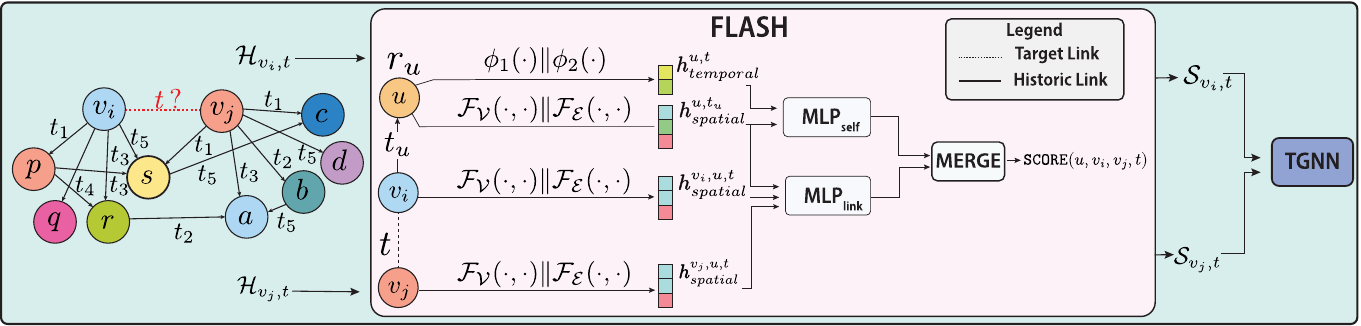}
    \caption{Overview of \ourmethod{}. Each historical neighbor $u$ is assigned a score based on its temporal, spatial, and structural relationships with $v_i$ and $v_j$. The highest-scoring neighbors are selected.}
    \label{fig:score}
\end{figure*}
For a given node $v$ and timestamp $t$, we denote the historic neighborhood $\gH_{v, t}$ as the set of all nodes that interacted with $v$ before time $t$:
\begin{align}
\label{eqn:H}
    \gH_{v, t} = \{u |(v,u)\in\gE \} 
\end{align}
Certain TGNNs allow the same node $u$ to appear multiple times in $\mathcal{H}_{v,t}$, where each occurrence of $u$ is associated with a distinct timestamp that records the time $v$ and $u$ interacted.

TGNNs construct node representations by sampling a subset $\mathcal{S}_{v, t}(k) \subseteq \mathcal{H}_{v, t}$ of $k$ historical neighbors. Existing approaches employ various static selection strategies:
\begin{align}
    \mathcal{S}^{tru}_{v, t}(k) = \{u | r_u \leq k \}
\end{align}
\begin{align}
    S^{\mathrm{uni}}_{v,t}(k)\sim \mathrm{Unif}\!\left\{\, S \subseteq \gH_{v,t} \;\middle|\; |S|=k \,\right\}.
\end{align}
where $r_u$ is the rank of $u$ with respect to the sequence of sorted neighbors from $\mathcal{H}_{v, t}$ in a decreasing order by their timestamp.
Some TGNNs~\cite{rossi2020temporal} allow sampling beyond the 1-hop historical neighborhood, either by allowing the sampling technique to sample farther nodes in advance or by applying the sampling technique recursively, i.e., sampling from the neighborhood of the sampled neighbors. 

Given the sampled neighborhood $\mathcal{S}_{v, t}(k)$, the representation of node $v$ at time $t$ is computed as:
\begin{align}
\label{eqn:z}
    \vz_v^t = \psi\left(\mathcal{S}_{v, t}(k)\right)
\end{align}
where $\psi(\cdot)$ is a parametric function of the TGNN that maps node to vector based on its sampled neighborhood. 

For the task of future link prediction, given a pair of nodes $(v_i, v_j)$ and their appropriate representations $\vz_{v_i}^t$ and $\vz_{v_j}^t$ at time $t$, a TGNN assigns a probability to the existence of a future edge between them using a learnable \texttt{MERGE} function:
\begin{align}
p(v_i, v_j \mid t)
     = \texttt{MERGE}(\vz_{v_i}^t, \vz_{v_j}^t)
\label{eqn:merge}
\end{align}
To train and evaluate TGNNs for future link prediction, the common approach is to split the entire sequence of interactions into two consecutive non-overlapping segments: a training prefix and an evaluation suffix. All interactions within the training prefix serve as positive examples, indicating node pairs that do form an edge at a specific time. For negative examples, random node pairs are sampled. The TGNN parameters are then updated by minimizing a binary classification loss (e.g., cross-entropy) that distinguishes positive from negative edges:
\begin{align}
\label{eqn:task}
\mathcal{L}_{\text{task}}
&= - \sum_{(v_i,v_j,t)\in\text{train}} \Big[
      y_{ij}^t \,\log p(v_i,v_j \vert t) \nonumber\\
&\quad +\; (1-y_{ij}^t)\,\log\,\bigl(1-p(v_i,v_j \vert t)\bigr) \Big]
\end{align}

where $y_{ij}^t$ is $1$ for observed edges in the training prefix and $0$ otherwise.
Once trained, the TGNN can be used to predict edges in the evaluation suffix by computing the probability $p(v_i, v_j \mid t)$ for new future node pairs.

\subsection{Theoretical analysis of Heuristic Neighborhood Samplings}
\label{subsec:expressivity_gap}

The lack of flexibility in common neighborhood sampling techniques, such as $k$ recent selection (truncation) or uniform sampling,  hinder the performance of TGNNs. Specifically, we show that there exist dynamic graphs such that any TGNN relying on these common heuristics cannot learn from them (i.e., it cannot overfit on this graph).

\begin{theorem}
\label{th:truncation_fails}
    For any $k$ there exists a dynamic graph on which a TGNN that applies $k$ recent neighbors sampling cannot learn.
\end{theorem}

\begin{theorem}
\label{th:uniform_fails}
    For any $k$, there exists a dynamic graph on which a TGNN that applies uniform sampling of $k$ historical neighbors cannot learn.
\end{theorem}

To prove \Cref{th:truncation_fails}, we find a graph that the $k+1$ recent neighbor is required to learn the dynamic behavior of the graph. To prove \Cref{th:uniform_fails}, we find a dynamic graph that requires consistently selecting the same recent neighbor to learn its dynamic behavior.
We provide the full proofs and assumptions for \Cref{th:truncation_fails} and \Cref{th:uniform_fails} in Appendix D.

\paragraph{Implication.}
These common sampling heuristics  discard potentially crucial historical interactions, limiting the performance of TGNNs and reduce their expressive power. We aim to develop a learnable and adaptive sampling technique that not only generalizes these heuristics but also enables TGNNs to become more expressive.

%% file: ijcai26/sections/Method.tex
\section{Method}
\label{sec:method}
In the previous section, we showed that current neighborhood sampling strategies do not fully account for the graph structure and its features, preventing TGNNs from capturing even simple evolving dynamics. Another limitation of these sampling heuristics stems from the fact that different TGNNs learn temporal dynamics in distinct ways due to their diverse designs. For example, TGN~\cite{rossi2020temporal} uses memory states, whereas DyGFormer~\cite{yu2023towards} employs a Transformer-based architecture~\cite{NIPS2017_3f5ee243} and jointly processes $\mathcal{S}{v_i, t}$ and $\mathcal{S}{v_j, t}$. Thus, enforcing the same fixed neighborhood sampling scheme across various TGNNs may undermine their performance. A learnable sampling strategy can adapt not only to the structure of the dynamic graph but also to the way each TGNN architecture exploits temporal signals.

\subsection{Desiderata for Adaptive Neighborhood Sampling in TGNNs}
We propose an adaptive neighborhood sampling mechanism that satisfies the following desiderata:
\begin{enumerate}[label=({\arabic*})]
    \item \textbf{Adapt to CTDG Dynamics:} We seek a mechanism $\texttt{SAMPLE}(\cdot)$ that accounts for node and edge attributes as well as their timestamps, i.e., the learnable parameters $\vtheta$ should be informed by the CTDG's interaction patterns. Formally, for a node $v$ with historical neighborhood $\mathcal{H}_{v,t}$ and a candidate interaction partner $v'$,
        \begin{align}
        \mathcal{S}_{v,t} = \texttt{SAMPLE}\!\left(v, v', \mathcal{H}_{v,t}; \vtheta, k\right),
        \end{align}
    where $|\mathcal{S}_{v,t}| = k$.

    \item \textbf{Generalize Existing Heuristics:} $\texttt{SAMPLE}(\cdot)$ should subsume commonly used heuristics (e.g., truncation, uniform sampling) as special cases via appropriate choices of $\vtheta$.

    \item \textbf{Seamless Integration with TGNNs:} Since TGNN backbones encode temporal dynamics differently (e.g., memory modules, attention), the relevance of sampled neighbors may depend on the backbone. Accordingly, the TGNN should inform the sampling procedure and the learning of $\vtheta$. Moreover, the mechanism should integrate seamlessly with a broad range of TGNNs, including models with non-differentiable feature extractors.

    \item \textbf{Self-Supervised Learning of Neighbor Importance:} Because ground-truth neighbor ``importance'' is unavailable, the mechanism should be learnable in a self-supervised manner that rewards neighbor subsets conducive to accurate link prediction.
\end{enumerate}

\subsection{\ourmethod{}}
To predict a future link $(v_i, v_j)$ at time $t$, we learn how informative each historical neighbor $u \in \mathcal{H}_{v_i,t}$ is for this prediction. Concretely, $\texttt{SAMPLE}(\cdot;\vtheta)$ assigns a score to each historical neighbor and selects the $k$ neighbors with the highest score.

\paragraph{Learning the scoring function.}
We first construct an informative representation for each historical neighbor that captures both \textbf{structural} (interaction context) and \textbf{temporal} (time-dependent) information (\textbf{D1}). For a neighbor $u \in \mathcal{H}_{v_i,t}$, let $t_u$ denote the time at which $u$ and $v_i$ interacted, and let $r_u$ denote the rank of $u$ when sorting $\mathcal{H}_{v_i,t}$ in decreasing order of interaction time. We define:

\begin{align}
    \vh^{u}_{\text{spatial}} &= \bigl[F_{\mathcal{V}}(u, t_u) \,\|\, F_{\mathcal{E}}(v_i, u, t_u) \,\|\, \gM(u)\bigr]\\
    \vh^{v_i,u,t}_{\text{spatial}} &= \bigl[F_{\mathcal{V}}(v_i, t_u) \,\|\, F_{\mathcal{V}}(v_i, t) \,\|\, \gM(v_i)\bigr]\\
    \vh^{v_j,u,t}_{\text{spatial}} &= \bigl[F_{\mathcal{V}}(v_j, t_u) \,\|\, F_{\mathcal{V}}(v_j, t) \,\|\, \gM(v_j)\bigr]\\
    \vh^{u,t}_{\text{temporal}} &= [\phi_1(t-t_u) \,\|\, \phi_2(r_u)],
\end{align}

where $\phi_1(\cdot)$ and $\phi_2(\cdot)$ are Time2Vec representations~\citep{kazemi2019time2vec}, $\gF_{\gV}$ maps nodes and timestamps to their corresponding features, $F_{\mathcal{E}}$ maps edges and timestamps to their corresponding features (as defined in \Cref{sec:background}), $\gM(\cdot)$ denotes learnable node features, and $\|$ denotes concatenation. Following the standard TGNN pipeline and prior work~\citep{deng2024taser}, we project the neighbor and link representations to a common dimensionality using MLPs, and then combine them via a $\texttt{MERGE}$ operator:

\begin{align}
\label{eqn:nomad}
\begin{aligned}
\texttt{SCORE}_{u,v_i,v_j,t} = \texttt{MERGE}\Big(
&\texttt{MLP}_{\texttt{self}}(\vh_{spatial}^u,\vh_{temporal}^{u,t}),\\
&\texttt{MLP}_{\texttt{link}}(\vh_{spatial}^u,\\
&\qquad\vh_{spatial}^{v_i,u,t},\vh_{spatial}^{v_j,u,t})
\Big)
\end{aligned}
\end{align}
where $\texttt{MERGE}$ is an MLP with a single hidden layer, as suggested for combining source--destination information~\citep{yu2023towards,tian2023freedyg}. $\texttt{SCORE}_{u,v_i,v_j,t}$ measures the utility of neighbor $u$ for predicting the link $(v_i,v_j)$ at time $t$, where larger values indicate higher importance. \ourmethod{} then selects the $k$ neighbors with the highest scores. \Cref{fig:score} provides a detailed schematic of \ourmethod{}.

\paragraph{Training \ourmethod{}.}
For each node pair $(v_i, v_j)$ at time $t$, let $y_{ij}^t \in \{0,1\}$ indicate whether $(v_i,v_j)$ is a \emph{positive} (observed) edge ($y_{ij}^t=1$) or a \emph{negative} (unobserved) edge ($y_{ij}^t=0$). Let a TGNN compute edge probabilities using the subsets selected by \ourmethod{},
$(\mathcal{S}_{v_i,v_j,t}, \mathcal{S}_{v_j,v_i,t})$, and using uniformly sampled subsets
$(\mathcal{S}_{v_i,t}^{\text{uni}}, \mathcal{S}_{v_j,t}^{\text{uni}})$. We define the probability difference:
\begin{align}
\Delta_{ij}^t 
&= p\!\left(v_i,v_j \vert t;\mathcal{S}_{v_i,v_j,t},\mathcal{S}_{v_j,v_i,t}\right) \notag\\
&\quad - p\!\left(v_i,v_j \vert t;\mathcal{S}_{v_i,t}^{\text{uni}},\mathcal{S}_{v_j,t}^{\text{uni}}\right).
\end{align}
Let $\overline{s_{v_i}},\overline{s_{v_j}}$ denote the average \ourmethod{} scores over nodes in
$\mathcal{S}_{v_i,v_j,t}$ and $\mathcal{S}_{v_j,v_i,t}$, respectively, and let
$\overline{s^{\text{uni}}_{v_i}},\overline{s^{\text{uni}}_{v_j}}$ denote the corresponding averages over
$\mathcal{S}_{v_i,t}^{\text{uni}}$ and $\mathcal{S}_{v_j,t}^{\text{uni}}$. For positive edges, we prefer $\Delta_{ij}^t>0$ (our subsets increase the predicted probability), while for negative edges we prefer $\Delta_{ij}^t<0$ (our subsets decrease it). We enforce this preference using a pairwise RankNet~\citep{burges2005learning} logistic loss:

\begin{align}
\label{eqn:loss}
\mathcal{L}_{ij}^t =
\begin{cases}
\begin{aligned}
&-\Bigl[\log\sigma\!\bigl(\overline{s_{v_j}}-\overline{s^{\text{uni}}_{v_j}}\bigr)\\
&\quad+\log\sigma\!\bigl(\overline{s_{v_i}}-\overline{s^{\text{uni}}_{v_i}}\bigr)\Bigr]
\end{aligned}
& \text{if } (y_{ij}^t-\tfrac12)\Delta_{ij}^t > 0, \\[6pt]
\begin{aligned}
&-\Bigl[\log\sigma\!\bigl(\overline{s^{\text{uni}}_{v_j}}-\overline{s_{v_j}}\bigr)\\
&\quad+\log\sigma\!\bigl(\overline{s^{\text{uni}}_{v_i}}-\overline{s_{v_i}}\bigr)\Bigr]
\end{aligned}
& \text{otherwise.}
\end{cases}
\end{align}

\input{ijcai26/sections/table1}
Minimizing $\mathcal{L}_{ij}^t$ encourages the score averages to move in the direction consistent with $y_{ij}^t$ without requiring any ground-truth importance supervision (\textbf{D3}). When the condition in \Cref{eqn:loss} holds, the neighborhoods selected by \ourmethod{} improve over uniform sampling, and the loss encourages higher scores for selected neighbors over the uniformly sampled neighbors. Otherwise, uniform sampling performs better and the loss correspondingly increases the scores of uniformly sampled neighbors relative to those selected by \ourmethod{}. We train \ourmethod{} jointly with a given TGNN by optimizing the sum of \Cref{eqn:loss} and \Cref{eqn:task}. Gradients from \Cref{eqn:task} update only the TGNN parameters, while for \Cref{eqn:loss} we stop gradients through $\Delta_{ij}^t$ so that updates flow only to \ourmethod{} parameters (\textbf{D4}). To ensure constant runtime, \ourmethod{} first applies a recent-neighbors finder~\citep{deng2024taser} to restrict the candidate set to a fixed size. It then computes scores only within this reduced neighborhood rather than over the full historical neighborhood. In Appendix G, we provide a throughout analysis of \ourmethod{} and TASER.

\subsection{Theoretical Analysis of \ourmethod{}}
The components of \ourmethod{} adapt to the evolution of the dynamic graph over time and to the interactions being predicted, by incorporating both spatiotemporal features and potentially interacting nodes when constructing each sampled neighborhood. We show that \ourmethod{} can replicate both $k$-recent neighbor selection and uniform sampling. Furthermore, we show that the graphs used in the proofs of \Cref{th:truncation_fails} and \Cref{th:uniform_fails} can be learned by a TGNN that utilizes \ourmethod{}. Therefore:
\begin{theorem}
\label{th:expressive}
\ourmethod{} is more expressive than $k$-recent neighbor selection and uniform sampling.
\end{theorem}
We provide the full proof of \Cref{th:expressive} in Appendix D.

%% file: ijcai26/sections/table1.tex
\begin{table*}[t]
    \centering
    \scriptsize
    \setlength{\tabcolsep}{3.0pt}
\renewcommand{\arraystretch}{1.0}
    
    \begin{tabular}{l c c c c c c c}
        \toprule
         \multirow{2}*{Method $\downarrow$ / Dataset $\rightarrow$} & \textsc{Wikipedia}  & \textsc{Reddit} & \textsc{Mooc} & \textsc{LastFM} & \textsc{Social Evo.}  & \textsc{Enron}  & \textsc{UCI}  \\
         & \textsc{AP} $\uparrow$ & \textsc{AP} $\uparrow$ & \textsc{AP} $\uparrow$ & \textsc{AP} $\uparrow$ & \textsc{AP} $\uparrow$ & \textsc{AP} $\uparrow$ & \textsc{AP} $\uparrow$\\
        \midrule
        \textsc{TGAT} + Trunc. & $94.05_{\pm0.06}$ & $93.63_{\pm0.16}$ & $79.69_{\pm0.24}$ & $65.64_{\pm0.34}$ & $85.36_{\pm0.25}$ & $72.91_{\pm0.58}$ & $79.74_{\pm0.37}$ \\
        \textsc{TGAT} + Uni. & $62.60_{\pm0.36}$ & $87.94_{\pm0.03}$ & $60.03_{\pm0.11}$ & $50.66_{\pm0.11}$ & $53.22_{\pm0.79}$ & $51.34_{\pm0.32}$ & $60.42_{\pm0.22}$ \\
        \textsc{TGAT} + NLB 
         & $91.49_{\pm0.25}$ & $92.78_{\pm0.11}$ & $77.37_{\pm0.17}$ & $65.64_{\pm0.34}$ & $83.19_{\pm0.07}$ & $70.69_{\pm0.93}$ & $77.80_{\pm1.25}$ \\ 
             \textsc{TGAT} + TASER & $67.78_{\pm0.99}$ & $89.14_{\pm0.59}$ & $67.10_{\pm0.56}$ & $52.28_{\pm0.52}$ & $70.40_{\pm0.45}$ & $60.55_{\pm0.91}$ & $70.62_{\pm0.33}$ \\
         \DashedMidrule
         \rowcolor{Wheat1!60}
         \textsc{TGAT} + \methodname{}  & $\mathbf{94.67}_{\pm0.38}$ & $\mathbf{95.92}_{\pm0.13}$ & $\mathbf{80.24}_{\pm0.54}$ & $\mathbf{74.94}_{\pm1.49}$ & $\mathbf{92.17}_{\pm0.25}$ & $\mathbf{79.04}_{\pm0.94}$ & $\mathbf{87.84}_{\pm0.12}$ \\
        \midrule
        \textsc{TGN} + Trunc. & $98.55_{\pm0.05}$ & $98.61_{\pm0.03}$ & $90.13_{\pm0.64}$ & $82.62_{\pm2.09}$ & $91.63_{\pm0.51}$ & $86.51_{\pm2.29}$ & $93.34_{\pm0.25}$ \\
        \textsc{TGN} + Uni. & $98.49_{\pm0.08}$ & $98.59_{\pm0.01}$ & $83.08_{\pm1.11}$ & $67.60_{\pm5.66}$ & $65.52_{\pm6.06}$ & $85.47_{\pm2.00}$ & $93.34_{\pm0.25}$ \\
        \textsc{TGN} + NLB & $98.31_{\pm0.09}$ & $98.61_{\pm0.04}$ & $89.81_{\pm0.60}$ & $80.45_{\pm1.94}$ & $91.19_{\pm0.38}$ & $86.51_{\pm2.29}$ & $92.91_{\pm0.43}$ \\
        \textsc{TGN} + TASER & $98.15_{\pm0.07}$ & $98.63_{\pm0.04}$ & $90.57_{\pm1.02}$ & $75.58_{\pm4.07}$ & $92.24_{\pm0.62}$ & $86.87_{\pm0.081}$ & $89.86_{\pm2.61}$ \\
         \DashedMidrule
        \rowcolor{Wheat1!60}
        \textsc{TGN} + \methodname{}  & $\mathbf{98.73}_{\pm0.06}$ & $\mathbf{99.06}_{\pm0.03}$ & $\mathbf{91.19}_{\pm0.51}$ & $\mathbf{90.54}_{\pm0.62}$ & $\mathbf{93.43}_{\pm0.11}$ & $\mathbf{89.90}_{\pm0.76}$ & $\mathbf{95.17}_{\pm0.16}$ \\
        
        \midrule
        \textsc{GraphMixer} + Trunc. & $96.23_{\pm0.24}$ & $95.17_{\pm0.03}$ & $80.71_{\pm0.11}$ & $72.98_{\pm0.07}$ & $87.09_{\pm0.12}$ & $81.63_{\pm0.47}$ & $93.14_{\pm0.44}$ \\
        \textsc{GraphMixer} + Uni. & $77.06_{\pm0.16}$ & $89.80_{\pm0.05}$ & $64.75_{\pm0.39}$ & $63.96_{\pm0.09}$ & $55.69_{\pm0.12}$ & $55.65_{\pm3.04}$ & $71.27_{\pm2.79}$ \\
        \textsc{GraphMixer} + NLB & $95.09_{\pm0.12}$ & $95.17_{\pm0.03}$ & $78.61_{\pm0.09}$ & $72.98_{\pm0.07}$ & $85.66_{\pm0.09}$ & $81.01_{\pm0.30}$ & $92.51_{\pm0.60}$ \\
                \textsc{GraphMixer} + TASER & $93.27_{\pm1.43}$ & $94.78_{\pm0.17}$ & $78.97_{\pm0.66}$ & $72.24_{\pm0.93}$ & $87.57_{\pm2.53}$ & $77.90_{\pm0.48}$ & $88.59_{\pm1.10}$ \\
        \DashedMidrule
        \rowcolor{Wheat1!60}
        \textsc{GraphMixer} + \methodname{}  & $\mathbf{97.51}_{\pm0.25}$ & $\mathbf{96.62}_{\pm0.11}$ & $\mathbf{80.85}_{\pm0.52}$ & $\mathbf{82.68}_{\pm0.74}$ & $\mathbf{92.84}_{\pm0.11}$ & $\mathbf{85.74}_{\pm0.70}$ & $\mathbf{93.21}_{\pm0.61}$ \\
         \midrule
        \textsc{DyGFormer} + Trunc. & $96.91_{\pm0.05}$ & $95.15_{\pm0.07}$ & $82.47_{\pm0.07}$ & $74.81_{\pm0.09}$ & $85.73_{\pm0.06}$ & $82.35_{\pm0.66}$ & $89.61_{\pm0.22}$ \\
        \textsc{DyGFormer} + Uni. & $96.87_{\pm0.07}$ & $95.15_{\pm0.07}$ & $82.47_{\pm0.07}$ & $74.81_{\pm0.09}$ & $85.71_{\pm0.06}$ & $82.35_{\pm0.66}$ & $89.61_{\pm0.22}$ \\
        \textsc{DyGFormer} + NLB & $96.74_{\pm0.07}$ & $95.03_{\pm0.09}$ & $81.09_{\pm0.05}$ & $74.52_{\pm0.18}$ & $85.40_{\pm0.10}$ & $81.86_{\pm0.45}$ & $89.12_{\pm0.19}$ \\
        \textsc{DyGFormer} + TASER & $95.47_{\pm0.32}$ & $93.83_{\pm0.33}$ & $81.56_{\pm0.67}$ & $72.72_{\pm1.73}$ & $90.09_{\pm0.11}$ & $81.74_{\pm0.64}$ & $88.15_{\pm0.59}$ \\
         \DashedMidrule
        \rowcolor{Wheat1!60}
        \textsc{DyGFormer} + \methodname{}  & $\mathbf{98.17}_{\pm0.04}$ & $\mathbf{98.11}_{\pm0.02}$ & $\mathbf{82.96}_{\pm0.42}$ & $\mathbf{86.09}_{\pm0.13}$ & $\mathbf{92.41}_{\pm0.11}$ & $\mathbf{88.70}_{\pm0.17}$ & $\mathbf{92.96}_{\pm0.16}$ \\
         \midrule
        \textsc{FreeDyG} + Trunc. & $98.35_{\pm0.01}$ & $97.53_{\pm0.02}$ & $85.02_{\pm0.02}$ & $80.19_{\pm0.04}$ & $90.88_{\pm0.05}$ & $88.77_{\pm0.20}$ & $95.21_{\pm0.33}$ \\
        \textsc{FreeDyG} + Uni. & $98.33_{\pm0.02}$ & $95.98_{\pm0.03}$ & $67.52_{\pm0.18}$ & $67.72_{\pm0.10}$ & $89.16_{\pm0.03}$ & $88.46_{\pm0.08}$ & $95.21_{\pm0.33}$
        \\
        \textsc{FreeDyG} + NLB & $98.33_{\pm0.01}$ & $97.59_{\pm0.02}$ & $83.56_{\pm0.12}$ & $80.02_{\pm0.09}$ & $90.40_{\pm0.05}$ & $88.77_{\pm0.20}$ & $95.17_{\pm0.26}$
        \\
        \textsc{FreeDyG} + TASER & $97.94_{\pm0.35}$ & $97.11_{\pm0.30}$ & $76.97_{\pm3.87}$ & $78.54_{\pm1.90}$ & $90.84_{\pm1.09}$ & $87.24_{\pm1.47}$ & $93.15_{\pm1.46}$ \\
        \DashedMidrule
        \rowcolor{Wheat1!60}
        \textsc{FreeDyG} + \methodname{}  & $\mathbf{98.94}_{\pm0.04}$ & $\mathbf{98.72}_{\pm0.02}$ & $\mathbf{85.69}_{\pm0.27}$ & $\mathbf{88.46}_{\pm0.07}$ & $\mathbf{93.62}_{\pm0.05}$ & $\mathbf{91.31}_{\pm0.01}$ & $\mathbf{95.86}_{\pm0.16}$ \\
        \bottomrule
    \end{tabular}
    \caption{Comparison of various neighborhood sampling methods on transductive future edge prediction using 2 historical neighbors on different datasets from DyGLib. The best performing method is in bold.}
    \label{tab:transductive-AP-Dyglib}
\end{table*}

%% file: ijcai26/sections/Experiments.tex
\section{Experiments}
\label{sec:experiments}

We evaluate \ourmethod{} across multiple dynamic graph benchmarks and compare it with recent state-of-the-art baselines. \Cref{exp:dyglib} presents our key empirical findings, and \Cref{exp:ablation} provides additional ablation studies. Additional results are reported in Appendix C. Our experiments aim to answer the following research questions:
\begin{itemize}[nosep]
\item \textbf{(RQ1)}~How does our proposed strategy compare to established sampling baselines in predictive accuracy?
\item \textbf{(RQ2)}~Does our method generalize across varying graph sizes, sparsity levels, and temporal granularities?

\item \textbf{(RQ3)}~What is the computational overhead of our approach relative to existing methods?
\end{itemize}
\input{ijcai26/sections/table2}

\paragraph{Experimental Setup.}
We test our method on five common TGNNs: \textsc{TGAT}~\cite{xu2020inductive}, \textsc{TGN}~\cite{rossi2020temporal}, \textsc{GraphMixer}~\cite{cong2023we}, \textsc{DyGFormer}~\cite{yu2023towards}, and \textsc{FreeDyG}~\cite{tian2023freedyg}. Detailed descriptions of each model are provided in Appendix B. Each model is trained with four neighbor sampling strategies: Truncation~\cite{cong2023we,yu2023towards}, Uniform~\cite{rossi2020temporal}, No-Look-Back (NLB)~\cite{luoscalable}, TASER with the recent-neighbors finder and a linear predictor~\cite{deng2024taser}, and our proposed method. Following standard practice, we use a chronological 70\%-15\%-15\% train-validation-test split. Additional parameters regarding the training scheme and implementation-specific details are provided in Appendix E.

\begin{figure}[t]
    \centering
    
    \includegraphics[width=\linewidth]{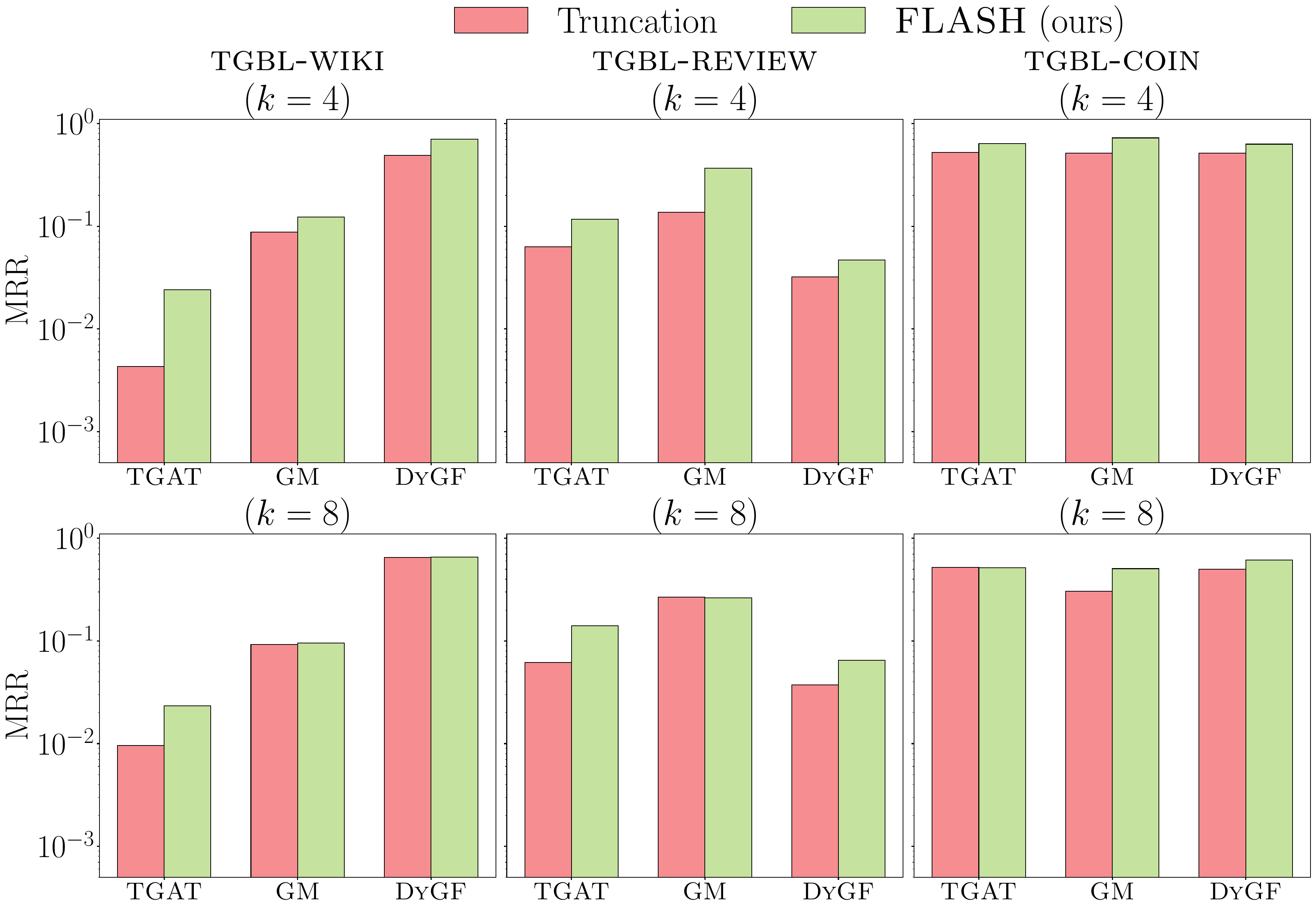}
    \caption{\methodname{} vs.\ Truncation on the TGB benchmark. Results are reported in MRR (dynamic link prediction) with random negative sampling over three runs, using $k=4$ and $k=8$ historical neighbors. Here \textsc{GM} refers to \textsc{GraphMixer} and \textsc{DyGF} refers to \textsc{DyGFormer}.}
    \label{fig:TGB}
\end{figure}

\begin{figure}[t]
    \centering
    \includegraphics[width=\linewidth]{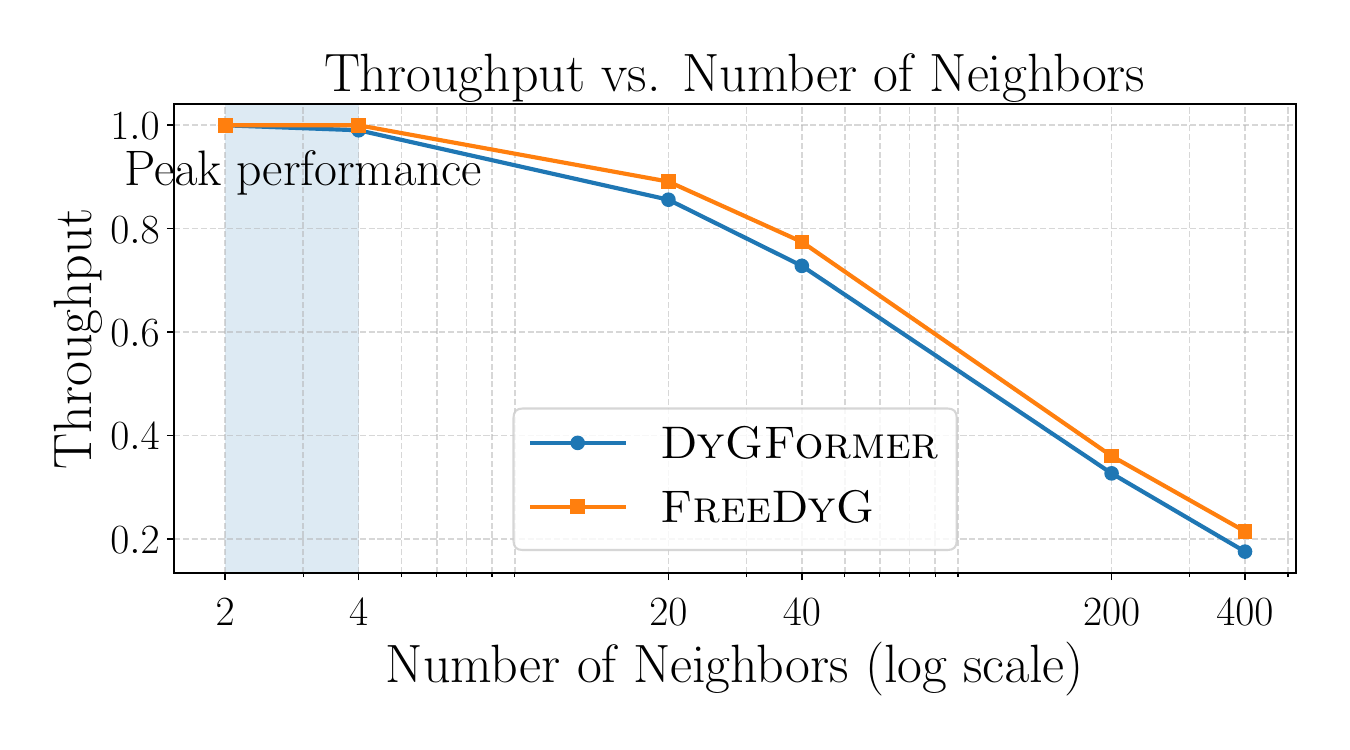}
    \caption{Throughput comparison of TGNNs as a function of selected neighborhood size, reported as the number of edges processed per second. Values are normalized by the throughput at neighborhood size 2 (e.g., if \textsc{DyGFormer} processes 200 edges/s with neighborhood size 2 and 100 edges/s with neighborhood size 40, the reported value is 0.5).
\label{tab:just}}
    \label{fig:just}
\end{figure}

\paragraph{Importance of Small Neighborhoods}
As discussed in \Cref{sec:introduction}, processing large neighborhoods is computationally prohibitive in both time and space. \Cref{fig:just} reports the throughput of state-of-the-art TGNNs for varying neighborhood sizes. The results show that increasing the neighborhood processed in parallel substantially degrades model throughput. Consequently, TGNNs should employ small neighborhoods (2–4) to achieve peak runtime performance. Accordingly, we evaluate neighborhood selection strategies under a small neighborhood budget (2–4) to determine which strategy is optimal for future link prediction while sustaining peak throughput in TGNNs.    

\subsection{Results}

\paragraph{Evaluation with DyGLib.}
\label{exp:dyglib}
We use the \textsc{Wikipedia}, \textsc{Reddit}, \textsc{Mooc}, \textsc{LastFM}, \textsc{Enron}, \textsc{Social Evo.}, and \textsc{UCI} datasets from the DyGLib benchmark~\cite{yu2023towards}. DyGLib includes datasets spanning a wide range of graph sizes, with some containing over a million edges (\textbf{RQ2}). Full dataset statistics are provided in Appendix A. We evaluate future edge prediction (\Cref{eqn:task}) under two settings: (1) \textit{transductive}, where all nodes appear during training, and (2) \textit{inductive}, where test nodes are unseen during training. \Cref{tab:transductive-AP-Dyglib,tab:inductive-AP-Dyglib} report results across models and methods. Overall, \methodname{} consistently outperforms previously suggested neighborhood selection strategies, both in the transductive setting and the inductive setting (\textbf{RQ1}). For example, on \textsc{Social Evo.}, \methodname{} achieves a \textbf{$\sim$8\%} relative improvement over truncation for \textsc{TGAT}, and on \textsc{LastFM}, a \textbf{$\sim$15\%} relative improvement for \textsc{FreeDyG}, demonstrating strong effectiveness even for computationally expensive models.

\paragraph{Evaluation with TGB.}
\label{exp:tgb}
We evaluate \methodname{} on the TGB benchmark~\cite{huang2024temporal} using \textsc{tgbl-wiki}, \textsc{tgbl-review}, and \textsc{tgbl-coin}. In contrast to DyGLib, TGB samples multiple random negative edges for each positive edge and is evaluated using Mean Reciprocal Rank (MRR). From \Cref{fig:TGB}, \methodname{} yields performance that is consistently on par with or better than baselines across TGNNs and neighborhood sizes. Notably, on \textsc{tgbl-review}, our method improves performance by $\sim$2.67$\times$ for \textsc{GraphMixer} when $k=4$, highlighting robustness across diverse dynamic graph scenarios (\textbf{RQ2}). 

\subsection{Ablation Study}
\label{exp:ablation}

\paragraph{\methodname{} Design Choices.}
We ablate key components of \methodname{}, namely the historical position embedding $\phi_2$, temporal embedding $\phi_1$, learnable node embedding $\gM$, and the link-awareness module $\texttt{MLP}_{link}$ in \Cref{eqn:nomad}, to assess the contribution of temporal, structural, and interaction context for future edge prediction (\textbf{RQ2}). \Cref{tab:ablation} shows that the positional embedding $\phi_2$ is particularly important on \textsc{Enron} and \textsc{UCI}, indicating that relying solely on temporal information is suboptimal. The interaction context module, $\texttt{MLP}_{link}$, is consistently beneficial across all datasets, providing reliable performance improvements.

\begin{table}[t]
\centering
\setlength{\tabcolsep}{1pt}
\begin{tabular}{lcccccc}
\toprule
$\phi_2$ & $\phi_1$ & $\gM$ & $\texttt{MLP}_{link}$ & \textsc{Wikipedia} & \textsc{Enron} & \textsc{UCI} \\
\midrule
\xmark & \cmark & \cmark & \cmark 
& 98.73$_{\pm0.07}$ 
& 89.63$_{\pm0.72}$ 
& 95.08$_{\pm0.19}$ \\
\xmark &\xmark & \cmark & \cmark 
& 98.20$_{\pm0.06}$ 
& 89.50$_{\pm0.74}$ 
& 95.08$_{\pm0.33}$ \\

\cmark & \cmark &\xmark & \cmark
& 98.58$_{\pm0.06}$ 
& 86.45$_{\pm1.25}$ 
& 93.27$_{\pm0.61}$ \\
\cmark & \cmark & \cmark &\xmark 
& 98.54$_{\pm0.09}$ 
& 87.38$_{\pm0.81}$ 
& 93.06$_{\pm0.60}$ \\
\midrule
\cmark & \cmark & \cmark & \cmark & \textbf{98.73}$_{\pm0.06}$ & \textbf{89.90}$_{\pm0.76}$ & \textbf{95.17}$_{\pm0.16}$\\
\bottomrule
\end{tabular}
\caption{Ablation of design choices of \methodname{}. (\cmark) denotes inclusion and (\xmark) denotes exclusion of the component. (\emph{transductive})}
\label{tab:ablation}
\label{tab:design}
\end{table} 
\paragraph{Computational Overhead.} \Cref{tab:runtimes} reports a throughput comparison for \methodname{}, measured as the number of processed edges per second, against the baselines \textbf{(RQ3)}. For a fair comparison, both TASER and \ourmethod{} employ the same neighbor finder, which constrains their neighborhoods to an identical size, this benefits both methods relative to uniform sampling. Because all experiments use a unified batch size, model latency is inversely proportional to throughput. Appendix~F further provides a time-complexity analysis of \methodname{} and the baselines.

%% file: ijcai26/sections/table2.tex
\begin{table*}[t]
    \centering
    \scriptsize
\setlength{\tabcolsep}{3.0pt}
    \renewcommand{\arraystretch}{1.0}
    \begin{tabular}{lccccccc}
        \toprule
         \multirow{2}*{Method $\downarrow$ / Dataset $\rightarrow$} & \textsc{Wikipedia}  & \textsc{Reddit} & \textsc{Mooc} & \textsc{LastFM} & \textsc{Social Evo.}  & \textsc{Enron}  & \textsc{UCI}  \\
         & \textsc{AP} $\uparrow$ & \textsc{AP} $\uparrow$ & \textsc{AP} $\uparrow$ & \textsc{AP} $\uparrow$ & \textsc{AP} $\uparrow$ & \textsc{AP} $\uparrow$ & \textsc{AP} $\uparrow$\\
        \midrule
         TGAT + Trunc. & $94.26_{\pm0.09}$ & $90.82_{\pm0.18}$ & $78.12_{\pm0.18}$ & $72.35_{\pm0.42}$ & $82.73_{\pm0.29}$ & $70.25_{\pm1.30}$ & $80.95_{\pm0.29}$ \\
         TGAT + Uni. & $62.77_{\pm0.43}$ & $82.85_{\pm0.13}$ & $58.38_{\pm0.14}$ & $50.75_{\pm0.07}$ & $53.06_{\pm0.40}$ & $51.63_{\pm0.72}$ & $59.65_{\pm0.36}$ \\
         TGAT + NLB. & $91.46_{\pm0.24}$ & $89.70_{\pm0.15}$ & $75.42_{\pm0.22}$ & $72.35_{\pm0.42}$ & $80.66_{\pm0.12}$ & $68.77_{\pm1.03}$ & $78.47_{\pm0.68}$ \\
      TGAT + TASER & $67.62_{\pm1.08}$ & $84.51_{\pm0.75}$ & $63.60_{\pm0.52}$ & $53.53_{\pm0.85}$ & $67.22_{\pm0.59}$ & $58.46_{\pm1.38}$ & $64.94_{\pm0.90}$ \\
                  \DashedMidrule
                  \rowcolor{Wheat1!60}
     TGAT+\methodname{} & $\mathbf{94.70}_{\pm0.35}$ & $\mathbf{94.24}_{\pm0.10}$ & $\mathbf{78.85}_{\pm0.61}$ & $\mathbf{79.56}_{\pm1.57}$ & $\mathbf{90.23}_{\pm0.25}$ & $\mathbf{76.70}_{\pm1.13}$ & $\mathbf{87.77}_{\pm0.17}$ \\
        \midrule
         TGN + Trunc. & $97.84_{\pm0.05}$ & $97.28_{\pm0.08}$ & $90.02_{\pm1.30}$ & $87.23_{\pm1.18}$ & $89.08_{\pm1.42}$ & $79.66_{\pm2.23}$ & $89.36_{\pm0.52}$ \\
         TGN + Uni. & $97.82_{\pm0.12}$ & $97.13_{\pm0.15}$ & $82.47_{\pm0.41}$ & $76.10_{\pm6.58}$ & $57.15_{\pm0.35}$ & $77.91_{\pm2.34}$ & $89.36_{\pm0.52}$  \\
         TGN + NLB. & $97.61_{\pm0.08}$ & $97.23_{\pm0.09}$ & $89.19_{\pm0.48}$ & $85.79_{\pm2.00}$ & $86.51_{\pm2.51}$ & $79.66_{\pm2.23}$ & $88.34_{\pm0.52}$ \\
        TGN + TASER & $97.39_{\pm0.17}$ & $97.33_{\pm0.12}$ & $89.90_{\pm1.18}$ & $81.97_{\pm3.73}$ & $90.31_{\pm1.50}$ & $78.99_{\pm2.56}$ & $85.28_{\pm1.22}$ \\
                  \DashedMidrule
        \rowcolor{Wheat1!60}
         TGN+\methodname{} & $\mathbf{98.08}_{\pm0.12}$ & $\mathbf{98.24}_{\pm0.07}$ & $\mathbf{90.50}_{\pm0.61}$ & $\mathbf{92.04}_{\pm0.40}$ & $\mathbf{92.06}_{\pm0.48}$ & $\mathbf{82.95}_{\pm1.02}$ & $\mathbf{92.40}_{\pm0.27}$ \\
        \midrule
        GraphMixer + Trunc. & $95.80_{\pm0.24}$ & $92.21_{\pm0.07}$ & $79.38_{\pm0.18}$ & $79.52_{\pm0.11}$ & $84.83_{\pm0.14}$ & $76.17_{\pm0.43}$ & $91.60_{\pm0.19}$ \\
         GraphMixer + Uni. & $78.72_{\pm0.12}$ & $84.29_{\pm0.10}$ & $65.91_{\pm0.30}$ & $73.74_{\pm0.07}$ & $51.13_{\pm0.15}$ & $52.46_{\pm3.33}$ & $71.69_{\pm1.43}$ \\
          GraphMixer + NLB & $94.34_{\pm0.10}$ & $92.21_{\pm0.07}$ & $76.89_{\pm0.10}$ & $79.52_{\pm0.11}$ & $83.09_{\pm0.08}$ & $74.99_{\pm0.34}$ & $90.89_{\pm0.23}$ \\
          GraphMixer +TASER & $92.42_{\pm1.54}$ & $91.25_{\pm0.30}$ & $76.99_{\pm0.59}$ & $79.43_{\pm1.02}$ & $85.10_{\pm2.73}$ & $70.70_{\pm0.44}$ & $86.78_{\pm1.76}$ \\
                   \DashedMidrule
        \rowcolor{Wheat1!60}
        GraphMixer + \methodname{} & $\mathbf{97.06}_{\pm0.29}$ & $\mathbf{94.77}_{\pm0.21}$ & $\mathbf{79.44}_{\pm0.55}$ & $\mathbf{86.70}_{\pm0.74}$ & $\mathbf{90.74}_{\pm0.11}$ & $\mathbf{80.61}_{\pm1.41}$ & $\mathbf{91.66}_{\pm0.24}$ \\
         \midrule
        DyGFormer + Trunc. & $97.02_{\pm0.07}$ & $93.40_{\pm0.08}$ & $81.27_{\pm0.09}$ & $79.97_{\pm0.11}$ & $82.86_{\pm0.07}$ & $80.46_{\pm0.91}$ & $89.99_{\pm0.18}$ \\
        DyGFormer + Uni. & $97.01_{\pm0.06}$ & $93.40_{\pm0.08}$ & $81.27_{\pm0.09}$ & $79.86_{\pm0.12}$ & $82.80_{\pm0.07}$ & $80.46_{\pm0.91}$ & $89.81_{\pm0.16}$ \\
        DyGFormer + NLB & $96.86_{\pm0.08}$ & $93.27_{\pm0.09}$ & $79.62_{\pm0.11}$ & $79.66_{\pm0.23}$ & $82.68_{\pm0.10}$ & $80.09_{\pm0.38}$ & $89.52_{\pm0.22}$ \\
        DyGFormer + TASER & $95.70_{\pm0.26}$ & $91.05_{\pm0.51}$ & $80.37_{\pm0.65}$ & $77.93_{\pm1.60}$ & $88.20_{\pm0.25}$ & $79.28_{\pm1.20}$ & $84.98_{\pm0.58}$ \\
                 \DashedMidrule
        \rowcolor{Wheat1!60}
        DyGFormer + \methodname{} & $\mathbf{97.91}_{\pm0.04}$ & $\mathbf{97.36}_{\pm0.04}$ & $\mathbf{81.42}_{\pm0.51}$ & $\mathbf{88.55}_{\pm0.15}$ & $\mathbf{90.56}_{\pm0.28}$ & $\mathbf{84.88}_{\pm0.66}$ & $\mathbf{92.73}_{\pm0.15}$ \\
         \midrule
        FreeDyG + Trunc. & $98.03_{\pm0.03}$ & $96.34_{\pm0.04}$ & $84.04_{\pm0.06}$ & $85.09_{\pm0.07}$ & $89.06_{\pm0.09}$ & $84.59_{\pm0.33}$ & $93.98_{\pm0.15}$ \\
        FreeDyG + Uni. & $98.01_{\pm0.02}$ & $93.86_{\pm0.04}$ & $68.53_{\pm0.24}$ & $76.41_{\pm0.09}$ & $86.92_{\pm0.09}$ & $83.96_{\pm0.21}$ & $93.98_{\pm0.15}$ \\
        FreeDyG + NLB. & $97.99_{\pm0.01}$ & $96.34_{\pm0.04}$ & $82.13_{\pm0.10}$ & $84.98_{\pm0.13}$ & $88.38_{\pm0.17}$ & $84.59_{\pm0.33}$ & $93.81_{\pm0.16}$ \\
        FreeDyG +TASER & $97.67_{\pm0.34}$ & $95.63_{\pm0.38}$ & $74.93_{\pm4.60}$ & $84.07_{\pm1.25}$ & $88.87_{\pm1.17}$ & $81.31_{\pm2.04}$ & $92.02_{\pm1.32}$ \\
                 \DashedMidrule
        \rowcolor{Wheat1!60}
        FreeDyG + \methodname{} & $\mathbf{98.59}_{\pm0.03}$ & $\mathbf{98.21}_{\pm0.04}$ & $\mathbf{84.38}_{\pm0.25}$ & $\mathbf{90.73}_{\pm0.09}$ & $\mathbf{91.53}_{\pm0.63}$ & $\mathbf{87.13}_{\pm0.62}$ & $\mathbf{94.60}_{\pm0.13}$ \\
        \bottomrule
    \end{tabular}
    \caption{Comparison of our suggested sampling strategy with other sampling strategies (Truncation, Uniform and NLB) in the inductive setting. Results are reported in AP for future edge prediction with random negative sampling over five different seeds. The significantly best result for each benchmark appears in bold font. 2 historical neighbors are used by each method.}
    \label{tab:inductive-AP-Dyglib}
\end{table*}

%% file: ijcai26/sections/Conclusion.tex
\section{Conclusion}
\label{sec:conclusion}

\begin{table}[t]
\centering
\setlength{\tabcolsep}{4.0pt}
\begin{tabular}{l cccc}
\toprule
Model & Uniform & NLB &  TASER & \methodname{} \\
\midrule
\textsc{TGAT} & 51\% & 44\% & 59\% & 75\% \\
\textsc{TGN}    & 67\% & 80\% & 82\% & 82\% \\
\textsc{GraphMixer} & 42\% & 40\% & 59\% &64\% \\
\textsc{DyGFormer}& 97\% & 87\% & 86\% &94\% \\
\textsc{FreeDyG}   & 84\% & 85\% & 91\% & 92\%\ \\
\bottomrule
\end{tabular}
\caption{Throughput comparison of neighbor sampling strategies. Values denote the number of edges processed per second by the TGNN relative to the Truncation baseline. For example, if TGAT with Uniform processes 300 edges/s over a full run while TGAT with Truncation processes 600 edges/s, the reported throughput for TGAT combined with Uniform is 50\%.}
\label{tab:runtimes}
\end{table}
In this work, we introduce \methodname{}, a flexible, end-to-end learnable neighborhood selection framework for dynamic graph learning. \methodname{} factorizes the neighbor scoring function into spatial and temporal components, ensuring that neighbor importance reflects both topological structure and temporal context. Our training objective couples the learned scores with downstream predictive performance, enabling effective optimization of the selection policy even when ground-truth importance scores are unavailable and TGNN operations are non-differentiable. This design allows \methodname{} to integrate seamlessly with a broad range of TGNN architectures, providing an efficient mechanism for selecting the most relevant historical neighbors. We further evaluate \methodname{} across multiple dynamic graph benchmarks and show that it matches or outperforms the considered baselines.

%% file: ijcai26/appendix/AppendixA.tex
\section{Dataset statistics and description}
\label{app:a}
In our empirical evaluation, we employed the following dynamic graph datasets, each capturing a distinct dynamic system and providing varied graph structures, edge features, and temporal resolutions:
\begin{itemize}

\item \textsc{Wikipedia }\citep{kumar2019predicting}:  
  This dataset contains one month of Wikipedia edit logs. Nodes represent editing users and Wikipedia pages, while edges represent individual edit requests. Each edge is timestamped and includes Linguistic Inquiry and Word Count (LIWC) \citep{pennebaker2001linguistic} feature vectors characterizing the textual content of the edit.

\item \textsc{Reddit} \citep{kumar2019predicting}:  
  This dataset comprises one month of Reddit posting logs. Nodes represent users and subreddits, and edges indicate posting actions. Each edge is timestamped to reflect the exact timing of a given post request, and equipped with LIWC features.

\item \textsc{Mooc} \citep{kumar2019predicting}:  
  This dataset records student interactions with Massive Open Online Courses (MOOCs). Nodes represent students and course content units (e.g., videos, assignments), and edges represent access actions such as viewing or submitting. Each edge is timestamped and is further annotated with four features describing the nature of the interaction.

\item \textsc{LastFM}~\cite{kumar2019predicting}:  
  Focusing on music listening behavior, this dataset tracks LastFM user activity over the course of a month. Nodes represent users and songs, and an edge between a user and a song signifies a listening event at a specific timestamp. No feature vectors are included in these edges.

\item \textsc{Enron}~\cite{shetty2004enron}:  
  This dataset consists of email exchange logs among Enron employees spanning three years. Each employee is modeled as a node, and each edge indicates a single email sent between two employees at a recorded timestamp. No additional edge features are provided.

\item \textsc{Social Evo.}~\cite{madan2011sensing}:  
  Derived from a study of undergraduate dormitory life over eight months, this dataset is presented as a proximity network of mobile phone interactions. Nodes represent individual participants, and edges capture observed proximities, each containing two distinct features describing the nature of the encounter.

\item \textsc{UCI}~\cite{panzarasa2009patterns}:  
  This dataset comprises a messaging log from an online student community at the University of California, Irvine. Each student is represented as a node, and an edge marks a message sent between two students, recorded with second-level granularity. No additional edge features are included beyond the timestamps.

\item \textsc{tgbl-wiki}~\cite{kumar2019predicting}: Based on the Wikipedia dataset, tgbl-wiki is a record of co-editing network on Wikipedia pages. This graph is bipartite, with editors and wiki pages serving as nodes, and each edge represents a user editing a page at a specific timestamp. Each edge also includes text features from the page edits. 

\item \textsc{tgbl-review}~\cite{ni2019justifying}: This dataset consists of an Amazon product review network spanning from 1997 to 2018, focusing on users and electronic products. Only users who submitted at least ten reviews during this period are retained in the dataset. Users rate these products on a five-point scale, forming a bipartite weighted graph in which users and products are the two sets of nodes in the bipartite graph. Each edge corresponds to a user’s review of a product at a specific time. 

\item \textsc{tgbl-coin}~\cite{shamsi2022chartalist}: This dataset captures cryptocurrency transactions based on the Stablecoin ERC20 transactions. Each node represents a cryptocurrency address, and each edge represents a fund transfer from one address to another at a specific time. The data spans from April 1, 2022 to November 1, 2022.
\end{itemize}

\begin{table*}[t]
    \centering
        \caption{Statistics of various datasets used in our experiments\label{fig:statistics}}
     \small    \begin{tabular}{l c c c c c c c}
     \toprule
     Dataset & Domain & \#Nodes & \#Edges & \#Node Features & \#Edge Features & Bipartite & Duration\\
     \midrule
    \textsc{Wikipedia}    & Social      & 9,227  & 157,474   &--& 172 & True  & 1 month       \\
    \textsc{Reddit}       & Social      & 10,984 & 672,447   &--& 172 & True  & 1 month       \\
   \textsc{Mooc}         & Interaction & 7,144  & 411,749   &--& 4   & True  & 17 months     \\
    \textsc{LastFM}       & Interaction & 1,980  & 1,293,103 &--& --   & True  & 1 month       \\
    \textsc{Enron }       & Social      & 184    & 125,235   &--& --   & False & 3 years       \\
    \textsc{Social Evo.}  & Proximity   & 74     & 2,099,519 &--& 2   & False & 8 months      \\
    \textsc{UCI}         & Social      & 1,899  & 59,835    &--& --  & False & 196 days      \\
    \textsc{tgbl-wiki}    & Social      & 9,227	& 157,474   &--& --  & True & 196 days      \\
    \textsc{tgbl-review}  & Social      & 352,637  & 4,873,540   &--& --   & True & 21 years      \\
    \textsc{tgbl-coin}    & Social      & 638,486  & 22,809,486    &--& --   & False & 7 months      \\
\bottomrule
    \end{tabular}
\end{table*}

%% file: ijcai26/appendix/AppendixB.tex
\section{Models description}
\label{app:B}
We employed five established temporal graph learning models in our experiments. A brief overview of each method is provided below:
\begin{itemize}

\item \textsc{TGAT}~\cite{xu2020inductive}:
\textsc{TGAT} employs a time-encoding function to capture continuous-time dynamics and uses self-attention to aggregate neighborhood information. The model computes node embeddings by jointly considering temporal features and the local structure of each node’s neighborhood.

\item \textsc{TGN}~\cite{rossi2020temporal}:
\textsc{TGN} introduces a general architecture for continuous-time dynamic graph (CTDG) tasks. It integrates two primary components: a prediction module and a memory module. The prediction module aggregates neighborhood information, while the memory module, implemented via an RNN, maintains up-to-date representations of node states. This design effectively addresses the staleness problem by considering neighborhood information.

\item \textsc{GraphMixer}~\cite{cong2023we}:
\textsc{GraphMixer} leverages three main components for future edge prediction. First, it uses an MLP-based link-encoder along with a fixed time-encoding function to process edge features. Second, a node-encoder applies neighborhood mean-pooling to capture contextual information for each node. Finally, a separate MLP is employed to predict the likelihood of future edges based on the encoded features.

\item \textsc{DyGFormer}~\cite{yu2023towards}:
\textsc{DyGFormer} is a transformer-based framework specifically designed for dynamic graph learning. It encodes each interaction by combining a co-occurrence embedding with a neighborhood representation for the interacting nodes. The model then applies a patching technique to historical representations of these nodes, thereby effectively capturing long-term temporal dependencies. These patches are passed through a transformer architecture, and their outputs are averaged to form the final interaction representation

\item \textsc{FreeDyG}~\cite{tian2023freedyg}: \textsc{FreeDyG} is an MLP-Mixer-based~\cite{tolstikhin2021mlp} architecture developed to effectively capture node interaction frequencies, thereby enhancing future edge prediction accuracy. The design comprises two core modules. First, the Node Interaction Frequency (NIF) Encoding augments co-neighborhood encoding with frequency-specific features. Second, a frequency-enhanced MLP-Mixer layer is introduced to efficiently capture periodic temporal patterns in the graph. By jointly modeling these frequency-sensitive components, \textsc{FreeDyG} aims to improve predictive performance for future interactions.

\end{itemize}

%% file: ijcai26/appendix/AppendixC.tex
\section{Additional results}
\label{app:additional_results}

In this section, we perform experiments in both \textit{transductive} and \textit{inductive} settings using 2 and 4 historical neighbors on DyGLib in 
\Cref{tab:transductive-AP-Dyglib-4},
\Cref{tab:transductive-ROC-Dyglib-2}, \Cref{tab:inductive-ROC-Dyglib-2}, 
\Cref{tab:transductive-ROC-Dyglib-4}, \Cref{tab:inductive-ROC-Dyglib-4}, \Cref{tab:inductive-AP-Dyglib-4}.

\begin{table*}[t]
    \centering
    \small
    \setlength{\tabcolsep}{2.5pt}
    \renewcommand{\arraystretch}{1.0}
    \caption{Comparison of various neighborhood sampling methods on \emph{transductive} future edge prediction using 4 historical neighbors on different datasets from DyGLib. Performance is measured in AP. The best performing method is marked in \textbf{bold}.}
    \begin{tabular}{l c c c c c c c}
        \toprule
         \multirow{2}*{Method $\downarrow$ / Dataset $\rightarrow$} & \textsc{Wikipedia}  & \textsc{Reddit} & \textsc{Mooc} & \textsc{LastFM} & \textsc{Social Evo.}  & \textsc{Enron}  & \textsc{UCI}  \\
         & \textsc{AP} $\uparrow$ & \textsc{AP} $\uparrow$ & \textsc{AP} $\uparrow$ & \textsc{AP} $\uparrow$ & \textsc{AP} $\uparrow$ & \textsc{AP} $\uparrow$ & \textsc{AP} $\uparrow$\\
        \midrule
        \textsc{TGAT} + Trunc. & $94.25_{\pm0.35}$ & $94.85_{\pm0.17}$ & $80.36_{\pm0.17}$ & $65.79_{\pm1.08}$ & $89.05_{\pm0.19}$ & $71.95_{\pm0.50}$ & $78.94_{\pm1.05}$ \\
        \textsc{TGAT} + Uni. & $69.74_{\pm0.35}$ & $92.30_{\pm0.04}$ & $61.30_{\pm0.03}$ & $50.79_{\pm0.08}$ & $57.00_{\pm0.07}$ & $52.39_{\pm0.19}$ & $65.55_{\pm0.46}$ \\
        \textsc{TGAT} + NLB & $91.22_{\pm0.31}$ & $94.41_{\pm0.32}$ & $77.79_{\pm0.19}$ & $63.98_{\pm0.97}$ & $86.06_{\pm0.18}$ & $69.13_{\pm1.23}$ & $77.88_{\pm0.88}$ \\ 
        \textsc{TGAT} + TASER & $75.19_{\pm0.26}$ & $93.01_{\pm0.40}$ & $69.09_{\pm0.55}$ & $53.93_{\pm0.99}$ & $74.75_{\pm0.71}$ & $60.90_{\pm2.57}$ & $73.00_{\pm1.02}$ \\
        \DashedMidrule
        \rowcolor{Wheat1!60}
        \textsc{TGAT} + \methodname{} & $\mathbf{95.07_{\pm0.55}}$ & $\mathbf{96.43_{\pm0.10}}$ & $\mathbf{80.59_{\pm0.37}}$ & $\mathbf{68.95_{\pm9.54}}$ & $\mathbf{92.26_{\pm0.12}}$ & $\mathbf{78.17_{\pm0.56}}$ & $\mathbf{79.75_{\pm1.61}}$ \\
        \midrule
        \textsc{TGN} + Trunc. & $98.50_{\pm0.08}$ & $98.61_{\pm0.03}$ & $90.09_{\pm1.26}$ & $79.25_{\pm2.10}$ & $92.50_{\pm0.44}$ & $87.18_{\pm1.32}$ & $93.19_{\pm0.46}$ \\
        \textsc{TGN} + Uni. & $96.86_{\pm0.24}$ & $98.63_{\pm0.04}$ & $83.92_{\pm0.96}$ & $68.87_{\pm3.05}$ & $72.20_{\pm8.27}$ & $85.29_{\pm1.68}$ & $88.78_{\pm1.98}$ \\
        \textsc{TGN} + NLB & $98.33_{\pm0.09}$ & $98.66_{\pm0.05}$ & $90.77_{\pm0.44}$ & $75.57_{\pm3.57}$ & $91.47_{\pm0.77}$ & $86.67_{\pm0.61}$ & $92.72_{\pm0.49}$ \\
        \textsc{TGN} + TASER & $98.27_{\pm0.10}$ & $98.65_{\pm0.01}$ & $90.39_{\pm0.88}$ & $75.66_{\pm2.39}$ & $92.92_{\pm0.85}$ & $86.26_{\pm0.47}$ & $92.59_{\pm0.44}$ \\
        \DashedMidrule
        \rowcolor{Wheat1!60}
        \textsc{TGN} + \methodname{} & $\mathbf{98.68_{\pm0.03}}$ & $\mathbf{99.00_{\pm0.02}}$ & $\mathbf{90.79_{\pm0.89}}$ & $\mathbf{86.82_{\pm0.91}}$ & $\mathbf{93.46_{\pm0.21}}$ & $\mathbf{89.09_{\pm1.33}}$ & $\mathbf{94.40_{\pm0.61}}$ \\
        \midrule
        \textsc{GraphMixer} + Trunc. & $96.42_{\pm0.06}$ & $96.06_{\pm0.20}$ & $81.29_{\pm0.22}$ & $74.63_{\pm0.19}$ & $90.30_{\pm0.08}$ & $81.52_{\pm0.15}$ & $\mathbf{92.86_{\pm0.57}}$ \\
        \textsc{GraphMixer} + Uni. & $82.28_{\pm0.33}$ & $93.56_{\pm0.06}$ & $67.64_{\pm0.14}$ & $64.71_{\pm0.20}$ & $56.84_{\pm0.29}$ & $57.85_{\pm0.89}$ & $71.33_{\pm2.41}$ \\
        \textsc{GraphMixer} + NLB & $95.66_{\pm0.05}$ & $96.42_{\pm0.02}$ & $79.69_{\pm0.06}$ & $74.57_{\pm0.10}$ & $88.75_{\pm0.04}$ & $81.76_{\pm0.26}$ & $91.64_{\pm0.50}$ \\
        \textsc{GraphMixer} + TASER & $95.27_{\pm0.79}$ & $96.12_{\pm0.10}$ & $80.21_{\pm2.73}$ & $74.23_{\pm0.27}$ & $89.57_{\pm1.16}$ & $79.51_{\pm0.53}$ & $90.71_{\pm0.61}$ \\
        \DashedMidrule
        \rowcolor{Wheat1!60}
        \textsc{GraphMixer} + \methodname{} & $\mathbf{97.16_{\pm0.32}}$ & $\mathbf{97.24_{\pm0.03}}$ & $\mathbf{81.52_{\pm0.22}}$ & $\mathbf{86.96_{\pm0.43}}$ & $\mathbf{93.52_{\pm0.07}}$ & $\mathbf{85.00_{\pm0.69}}$ & $92.62_{\pm0.58}$ \\
        \midrule
        \textsc{DyGFormer} + Trunc. & $98.10_{\pm0.03}$ & $97.52_{\pm0.04}$ & $85.34_{\pm0.08}$ & $79.28_{\pm0.16}$ & $92.93_{\pm0.01}$ & $85.15_{\pm0.60}$ & $93.53_{\pm0.16}$ \\
        \textsc{DyGFormer} + Uni. & $98.08_{\pm0.04}$ & $97.48_{\pm0.03}$ & $85.32_{\pm0.13}$ & $79.34_{\pm0.17}$ & $92.94_{\pm0.04}$ & $85.42_{\pm0.14}$ & $93.15_{\pm0.10}$ \\
        \textsc{DyGFormer} + NLB & $98.18_{\pm0.04}$ & $97.56_{\pm0.05}$ & $83.52_{\pm0.18}$ & $79.14_{\pm0.09}$ & $92.31_{\pm0.05}$ & $86.45_{\pm0.26}$ & $92.72_{\pm0.19}$ \\
        \textsc{DyGFormer} + TASER & $97.27_{\pm0.34}$ & $97.09_{\pm0.24}$ & $84.49_{\pm0.34}$ & $78.42_{\pm0.86}$ & $92.88_{\pm0.27}$ & $84.76_{\pm0.14}$ & $92.35_{\pm0.35}$ \\
        \DashedMidrule
        \rowcolor{Wheat1!60}
        \textsc{DyGFormer} + \methodname{} & $\mathbf{98.73_{\pm0.02}}$ & $\mathbf{98.71_{\pm0.01}}$ & $\mathbf{86.03_{\pm0.16}}$ & $\mathbf{87.37_{\pm0.03}}$ & $\mathbf{93.74_{\pm0.04}}$ & $\mathbf{89.84_{\pm0.23}}$ & $\mathbf{94.60_{\pm0.11}}$ \\
        \midrule
        \textsc{FreeDyG} + Trunc. & $98.73_{\pm0.03}$ & $98.27_{\pm0.01}$ & $\mathbf{86.79_{\pm0.04}}$ & $83.19_{\pm0.06}$ & $93.31_{\pm0.02}$ & $89.50_{\pm0.12}$ & $95.68_{\pm0.11}$ \\
        \textsc{FreeDyG} + Uni. & $97.52_{\pm0.04}$ & $97.64_{\pm0.04}$ & $71.28_{\pm0.12}$ & $70.67_{\pm0.13}$ & $75.26_{\pm0.05}$ & \textsc{n/a} & $87.25_{\pm0.13}$ \\
        \textsc{FreeDyG} + NLB & $98.78_{\pm0.02}$ & $98.38_{\pm0.01}$ & \textsc{n/a} & $83.17_{\pm0.05}$ & $92.88_{\pm0.01}$ & $89.93_{\pm0.20}$ & $96.13_{\pm0.18}$ \\
        \textsc{FreeDyG} + TASER & $98.61_{\pm0.23}$ & $98.21_{\pm0.09}$ & $82.98_{\pm2.63}$ & $82.99_{\pm0.52}$ & $92.65_{\pm0.28}$ & $89.51_{\pm0.61}$ & $94.46_{\pm0.49}$ \\

        \DashedMidrule
        \rowcolor{Wheat1!60}
        \textsc{FreeDyG} + \methodname{} & $\mathbf{99.11_{\pm0.02}}$ & $\mathbf{98.96_{\pm0.01}}$ & $86.59_{\pm0.16}$ & $\mathbf{88.95_{\pm0.13}}$ & $\mathbf{93.97_{\pm0.11}}$ & $\mathbf{91.66_{\pm0.16}}$ & $\mathbf{96.43_{\pm0.18}}$ \\
        \bottomrule
    \end{tabular}
    \label{tab:transductive-AP-Dyglib-4}
\end{table*}

\begin{table*}[t]
    \centering
    \small
    \setlength{\tabcolsep}{0.3pt}
    \renewcommand{\arraystretch}{1.0}
    \caption{Comparison of various neighborhood sampling methods on \emph{transductive} future edge prediction using 2 historical neighbors on different datasets from DyGLib. Performance is measured in ROC-AUC. The best performing method is marked in \textbf{bold}.}
    \begin{tabular}{l c c c c c c c}
        \toprule
         \multirow{2}*{Method $\downarrow$ / Dataset $\rightarrow$} & \textsc{Wikipedia}  & \textsc{Reddit} & \textsc{Mooc} & \textsc{LastFM} & \textsc{Social Evo.}  & \textsc{Enron}  & \textsc{UCI}  \\
         & \textsc{ROC-AUC} $\uparrow$ & \textsc{ROC-AUC} $\uparrow$ & \textsc{ROC-AUC} $\uparrow$ & \textsc{ROC-AUC} $\uparrow$ & \textsc{ROC-AUC} $\uparrow$ & \textsc{ROC-AUC} $\uparrow$ & \textsc{ROC-AUC} $\uparrow$\\
        \midrule
        \textsc{TGAT} + Trunc. & $92.80_{\pm0.07}$ & $92.88_{\pm0.19}$ & $80.60_{\pm0.20}$ & $62.93_{\pm2.70}$ & $88.13_{\pm0.16}$ & $68.05_{\pm0.89}$ & $75.53_{\pm0.38}$ \\
        \textsc{TGAT} + Uni. & $60.97_{\pm0.16}$ & $87.40_{\pm0.02}$ & $63.23_{\pm0.05}$ & $49.97_{\pm0.13}$ & $53.05_{\pm0.87}$ & $51.28_{\pm0.30}$ & $62.69_{\pm0.26}$ \\
        \textsc{TGAT} + NLB & $89.92_{\pm0.29}$ & $91.96_{\pm0.12}$ & $78.37_{\pm0.15}$ & $63.57_{\pm0.56}$ & $85.83_{\pm0.11}$ & $66.92_{\pm1.06}$ & $75.02_{\pm1.23}$ \\
                        \textsc{TGAT} + TASER & $64.76_{\pm1.06}$ & $88.33_{\pm0.57}$ & $68.92_{\pm0.51}$ & $51.59_{\pm0.55}$ & $72.50_{\pm0.56}$ & $58.21_{\pm0.84}$ & $73.50_{\pm0.43}$ \\
        \DashedMidrule
        \rowcolor{Wheat1!60}
        \textsc{TGAT} + \methodname{} & $\mathbf{93.36_{\pm0.27}}$ & $\mathbf{95.30_{\pm0.14}}$ & $\mathbf{81.02_{\pm0.73}}$ & $\mathbf{69.20_{\pm1.70}}$ & $\mathbf{93.98_{\pm0.32}}$ & $\mathbf{74.15_{\pm1.20}}$ & $\mathbf{83.79_{\pm0.24}}$ \\

        \midrule
        \textsc{TGN} + Trunc. & $98.48_{\pm0.05}$ & $98.58_{\pm0.02}$ & $91.92_{\pm0.82}$ & $82.26_{\pm2.08}$ & $93.89_{\pm0.38}$ & $87.78_{\pm0.56}$ & $92.75_{\pm0.22}$ \\
        \textsc{TGN} + Uni. & $98.39_{\pm0.08}$ & $98.56_{\pm0.01}$ & $86.32_{\pm0.96}$ & $68.21_{\pm5.12}$ & $71.63_{\pm5.48}$ & $87.51_{\pm2.09}$ & $92.98_{\pm0.29}$ \\
        \textsc{TGN} + NLB & $98.23_{\pm0.09}$ & $98.59_{\pm0.04}$ & $91.71_{\pm0.47}$ & $80.33_{\pm1.76}$ & $93.48_{\pm0.38}$ & $88.72_{\pm2.41}$ & $92.75_{\pm0.35}$ \\

\textsc{TGN} + TASER & $98.07_{\pm0.07}$ & $98.61_{\pm0.03}$ & $92.38_{\pm0.70}$ & $76.45_{\pm3.60}$ & $94.38_{\pm0.53}$ & $78.99_{\pm2.56}$ & $89.91_{\pm2.23}$ \\
        \DashedMidrule
        \rowcolor{Wheat1!60}
        \textsc{TGN} + \methodname{} & $\mathbf{98.62_{\pm0.08}}$ & $\mathbf{99.01_{\pm0.03}}$ & $\mathbf{92.75_{\pm0.45}}$ & $\mathbf{88.79_{\pm0.85}}$ & $\mathbf{95.44_{\pm0.13}}$ & $\mathbf{91.15_{\pm0.45}}$ & $\mathbf{94.51_{\pm0.19}}$ \\
        \midrule
        \textsc{GraphMixer} + Trunc. & $95.76_{\pm0.27}$ & $94.65_{\pm0.05}$ & $81.92_{\pm0.11}$ & $70.66_{\pm1.74}$ & $89.61_{\pm0.07}$ & $83.55_{\pm0.24}$ & $91.01_{\pm0.57}$ \\
        \textsc{GraphMixer} + Uni. & $74.98_{\pm0.24}$ & $89.47_{\pm0.05}$ & $68.54_{\pm0.37}$ & $61.81_{\pm0.20}$ & $56.94_{\pm0.16}$ & $55.91_{\pm2.95}$ & $68.96_{\pm2.88}$ \\
        \textsc{GraphMixer} + NLB & $94.61_{\pm0.09}$ & $94.83_{\pm0.03}$ & $79.89_{\pm0.07}$ & $71.07_{\pm0.12}$ & $88.10_{\pm0.08}$ & $83.60_{\pm0.15}$ & $90.23_{\pm0.79}$ \\
                        \textsc{GraphMixer} + TASER & $92.53_{\pm1.58}$ & $94.48_{\pm0.17}$ & $80.24_{\pm0.55}$ & $70.54_{\pm0.83}$ & $89.62_{\pm2.42}$ & $81.33_{\pm0.37}$ & $85.59_{\pm1.48}$ \\

        \DashedMidrule
        \rowcolor{Wheat1!60}
        \textsc{GraphMixer} + \methodname{} & $\mathbf{97.08_{\pm0.28}}$ & $\mathbf{96.32_{\pm0.12}}$ & $\mathbf{81.95_{\pm0.44}}$ & $\mathbf{79.30_{\pm1.01}}$ & $\mathbf{95.08_{\pm0.04}}$ & $\mathbf{86.07_{\pm0.70}}$ & $\mathbf{91.07_{\pm0.86}}$ \\
        \midrule
        \textsc{DyGFormer} + Trunc. & $96.00_{\pm0.08}$ & $93.87_{\pm0.03}$ & $82.23_{\pm0.04}$ & $69.88_{\pm0.16}$ & $87.04_{\pm0.06}$ & $78.41_{\pm1.02}$ & $84.94_{\pm0.12}$ \\
        \textsc{DyGFormer} + Uni. & $95.95_{\pm0.10}$ & $93.88_{\pm0.10}$ & $82.24_{\pm0.08}$ & $70.06_{\pm0.16}$ & $86.98_{\pm0.06}$ & $79.02_{\pm0.95}$ & $85.08_{\pm0.44}$ \\
        \textsc{DyGFormer} + NLB & $95.72_{\pm0.11}$ & $93.70_{\pm0.15}$ & $80.52_{\pm0.07}$ & $69.74_{\pm0.23}$ & $86.67_{\pm0.10}$ & $78.30_{\pm0.77}$ & $84.38_{\pm0.24}$ \\

\textsc{DyGFormer} + TASER & $94.15_{\pm0.36}$ & $92.43_{\pm0.37}$ & $81.35_{\pm0.73}$ & $66.54_{\pm1.81}$ & $91.57_{\pm0.22}$ & $78.49_{\pm0.93}$ & $86.02_{\pm0.54}$ \\
        \DashedMidrule
        \rowcolor{Wheat1!60}
        \textsc{DyGFormer} + \methodname{} & $\mathbf{97.67_{\pm0.06}}$ & $\mathbf{97.62_{\pm0.02}}$ & $\mathbf{82.85_{\pm0.51}}$ & $\mathbf{82.83_{\pm0.21}}$ & $\mathbf{94.61_{\pm0.13}}$ & $\mathbf{87.69_{\pm0.14}}$ & $\mathbf{89.68_{\pm0.23}}$ \\
        \midrule
        \textsc{FreeDyG} + Trunc. & $98.11_{\pm0.03}$ & $96.89_{\pm0.07}$ & $84.94_{\pm0.03}$ & $76.61_{\pm0.08}$ & $92.71_{\pm0.03}$ & $89.33_{\pm0.10}$ & $93.22_{\pm0.32}$ \\
        \textsc{FreeDyG} + Uni. & $98.12_{\pm0.01}$ & $95.17_{\pm0.05}$ & $70.30_{\pm0.23}$ & $63.73_{\pm0.17}$ & $90.54_{\pm0.03}$ & $89.46_{\pm0.13}$ & $93.51_{\pm0.57}$ \\
        \textsc{FreeDyG} + NLB & $98.09_{\pm0.02}$ & $97.13_{\pm0.02}$ & $83.22_{\pm0.12}$ & $76.44_{\pm0.10}$ & $91.98_{\pm0.05}$ & $90.02_{\pm0.19}$ & $93.44_{\pm0.44}$ \\
\textsc{FreeDyG} + TASER & $97.49_{\pm0.51}$ & $96.67_{\pm0.33}$ & $77.20_{\pm3.58}$ & $75.51_{\pm1.87}$ & $92.56_{\pm1.17}$ & $87.95_{\pm1.67}$ & $90.78_{\pm2.03}$ \\
        \DashedMidrule
        \rowcolor{Wheat1!60}
        \textsc{FreeDyG} + \methodname{} & $\mathbf{98.82_{\pm0.05}}$ & $\mathbf{98.51_{\pm0.03}}$ & $\mathbf{85.47_{\pm0.28}}$ & $\mathbf{86.29_{\pm0.06}}$ & $\mathbf{95.76_{\pm0.03}}$ & $\mathbf{91.96_{\pm0.32}}$ & $\mathbf{94.32_{\pm0.30}}$ \\
        \bottomrule
    \end{tabular}
    \label{tab:transductive-ROC-Dyglib-2}
\end{table*}

\begin{table*}[t]
    \centering
    \small
    \setlength{\tabcolsep}{2.5pt}
    \renewcommand{\arraystretch}{1.0}
    \caption{Comparison of various neighborhood sampling methods on \emph{inductive} future edge prediction using 2 historical neighbors on different datasets from DyGLib. Performance is measured in AP. The best performing method is marked in \textbf{bold}.}
    \begin{tabular}{l c c c c c c c}
        \toprule
         \multirow{2}*{Method $\downarrow$ / Dataset $\rightarrow$} & \textsc{Wikipedia}  & \textsc{Reddit} & \textsc{Mooc} & \textsc{LastFM} & \textsc{Social Evo.}  & \textsc{Enron}  & \textsc{UCI}  \\
         & \textsc{AP} $\uparrow$ & \textsc{AP} $\uparrow$ & \textsc{AP} $\uparrow$ & \textsc{AP} $\uparrow$ & \textsc{AP} $\uparrow$ & \textsc{AP} $\uparrow$ & \textsc{AP} $\uparrow$\\
        \midrule
        \textsc{TGAT} + Trunc. & $93.25_{\pm0.11}$ & $89.98_{\pm0.19}$ & $78.98_{\pm0.18}$ & $68.96_{\pm4.85}$ & $86.09_{\pm0.17}$ & $66.47_{\pm1.89}$ & $76.86_{\pm0.50}$  \\
        \textsc{TGAT} + Uni. & $60.92_{\pm0.36}$ & $82.34_{\pm0.16}$ & $60.60_{\pm0.12}$ & $50.11_{\pm0.06}$ & $52.27_{\pm0.95}$ & $50.52_{\pm1.30}$ & $60.90_{\pm0.32}$ \\
        \textsc{TGAT} + NLB & $90.18_{\pm0.27}$ & $88.87_{\pm0.23}$ & $76.28_{\pm0.19}$ & $70.44_{\pm0.66}$ & $83.41_{\pm0.16}$ & $65.17_{\pm1.31}$ & $75.33_{\pm0.82}$ \\ 
                        \textsc{TGAT} + \textsc{TASER} & $64.59_{\pm1.21}$ & $83.70_{\pm0.70}$ & $65.19_{\pm0.56}$ & $52.95_{\pm1.03}$ & $69.01_{\pm0.76}$ & $55.75_{\pm1.29}$ & $65.35_{\pm0.50}$ \\
        \DashedMidrule
        \rowcolor{Wheat1!60}
        \textsc{TGAT} + \methodname{} &  $\mathbf{93.60_{\pm0.21}}$ & $\mathbf{93.43_{\pm0.08}}$ & $\mathbf{79.53_{\pm0.83}}$ & $\mathbf{74.55_{\pm2.02}}$ & $\mathbf{92.43_{\pm0.32}}$ & $\mathbf{72.05_{\pm1.34}}$ & $\mathbf{83.81_{\pm0.26}}$\\
        \midrule
        \textsc{TGN} + Trunc. & $97.73_{\pm0.09}$ & $97.17_{\pm0.07}$ & $91.65_{\pm1.18}$ & $86.93_{\pm1.07}$ & $91.74_{\pm1.15}$ & $78.20_{\pm1.88}$ & $87.53_{\pm0.63}$ \\
        \textsc{TGN} + Uni. & $97.69_{\pm0.13}$ & $97.03_{\pm0.17}$ & $85.48_{\pm0.42}$ & $76.33_{\pm5.87}$ & $60.10_{\pm2.36}$ & $78.46_{\pm2.31}$ & $87.94_{\pm0.51}$ \\
        \textsc{TGN} + NLB & $97.46_{\pm0.08}$ & $97.15_{\pm0.07}$ & $91.15_{\pm0.35}$ & $85.53_{\pm1.87}$ & $89.18_{\pm2.12}$ & $81.42_{\pm1.96}$ & $87.08_{\pm0.72}$ \\

        \textsc{TGN} + \textsc{TASER} & $97.22_{\pm0.20}$ & $97.26_{\pm0.14}$ & $91.69_{\pm0.90}$ & $82.71_{\pm2.77}$ & $92.52_{\pm0.99}$ & $80.68_{\pm3.08}$ & $83.48_{\pm1.06}$ \\
        \DashedMidrule
        \rowcolor{Wheat1!60}
        \textsc{TGN} + \methodname{}  & $\mathbf{97.95_{\pm0.15}}$ & $\mathbf{98.07_{\pm0.07}}$ & $\mathbf{92.00_{\pm0.50}}$ & $\mathbf{90.81_{\pm0.61}}$ & $\mathbf{94.30_{\pm0.28}}$ & $\mathbf{83.32_{\pm0.65}}$ & $\mathbf{90.78_{\pm0.40}}$ \\
        \midrule
        \textsc{GraphMixer} + Trunc. & $95.28_{\pm0.23}$ & $91.69_{\pm0.10}$ & $\mathbf{80.61_{\pm0.17}}$ & $77.43_{\pm1.66}$ & $87.76_{\pm0.10}$ & $76.44_{\pm0.27}$ & $89.83_{\pm0.26}$ \\
        \textsc{GraphMixer} + Uni. & $76.13_{\pm0.17}$ & $83.91_{\pm0.08}$ & $68.72_{\pm0.36}$ & $70.83_{\pm0.12}$ & $50.14_{\pm0.33}$ & $50.64_{\pm4.12}$ & $69.10_{\pm1.36}$   \\
        \textsc{GraphMixer} + NLB & $93.71_{\pm0.11}$ & $91.64_{\pm0.06}$ & $78.14_{\pm0.13}$ & $77.87_{\pm0.11}$ & $85.77_{\pm0.10}$ & $75.80_{\pm0.46}$ & $88.96_{\pm0.29}$ \\
                        \textsc{GraphMixer} + \textsc{TASER} & $91.55_{\pm1.66}$ & $90.75_{\pm0.29}$ & $78.15_{\pm0.51}$ & $77.88_{\pm0.98}$ & $87.19_{\pm2.71}$ & $71.66_{\pm0.41}$ & $83.94_{\pm2.01}$ \\

        \DashedMidrule
        \rowcolor{Wheat1!60}
        \textsc{GraphMixer} + \methodname{} & $\mathbf{96.59_{\pm0.36}}$ & $\mathbf{94.36_{\pm0.26}}$ & $80.57_{\pm0.46}$ & $\mathbf{84.30_{\pm1.00}}$ & $\mathbf{93.62_{\pm0.11}}$ & $\mathbf{79.06_{\pm1.54}}$ & $\mathbf{89.86_{\pm0.28}}$ \\
        \midrule
        \textsc{DyGFormer} + Trunc. & $96.34_{\pm0.07}$ & $91.56_{\pm0.02}$ & $80.89_{\pm0.05}$ & $75.90_{\pm0.19}$ & $84.22_{\pm0.06}$ & $76.36_{\pm1.16}$ & $85.99_{\pm0.21}$ \\
        \textsc{DyGFormer} + Uni. & $96.31_{\pm0.06}$ & $91.61_{\pm0.15}$ & $80.89_{\pm0.10}$ & $76.06_{\pm0.20}$ & $84.16_{\pm0.05}$ & $77.23_{\pm1.62}$ & $85.96_{\pm0.27}$ \\
        \textsc{DyGFormer} + NLB & $96.09_{\pm0.12}$ & $91.35_{\pm0.15}$ & $78.80_{\pm0.13}$ & $75.82_{\pm0.25}$ & $83.84_{\pm0.09}$ & $76.75_{\pm0.87}$ & $85.54_{\pm0.25}$ \\

        \textsc{DyGFormer} + \textsc{TASER} & $94.71_{\pm0.33}$ & $88.77_{\pm0.54}$ & $79.93_{\pm0.77}$ & $72.35_{\pm1.81}$ & $88.20_{\pm0.25}$ & $76.02_{\pm1.37}$ & $81.08_{\pm0.56}$ \\
        \DashedMidrule
        \rowcolor{Wheat1!60}
        \textsc{DyGFormer} + \methodname{} & $\mathbf{97.48_{\pm0.05}}$ & $\mathbf{96.68_{\pm0.04}}$ & $\mathbf{81.11_{\pm0.72}}$ & $\mathbf{85.94_{\pm0.23}}$ & $\mathbf{93.13_{\pm0.27}}$ & $\mathbf{84.12_{\pm0.91}}$ & $\mathbf{89.91_{\pm0.21}}$ \\
        \midrule
        \textsc{FreeDyG} + Trunc. & $97.71_{\pm0.04}$ & $95.31_{\pm0.09}$ & $83.92_{\pm0.04}$ & $82.38_{\pm0.09}$ & $91.18_{\pm0.02}$ & $83.96_{\pm0.61}$ & $91.84_{\pm0.18}$ \\
        \textsc{FreeDyG} + Uni. & $97.70_{\pm0.02}$ & $92.45_{\pm0.04}$ & $70.53_{\pm0.32}$ & $72.53_{\pm0.13}$ & $88.20_{\pm0.09}$ & $84.17_{\pm0.44}$ & $92.06_{\pm0.22}$ \\
        \textsc{FreeDyG} + NLB & $97.66_{\pm0.02}$ & $95.59_{\pm0.05}$ & $81.66_{\pm0.15}$ & $82.30_{\pm0.10}$ & $90.19_{\pm0.14}$ & $85.06_{\pm0.33}$ & $91.79_{\pm0.13}$ \\

        \textsc{FreeDyG} + \textsc{TASER} & $97.20_{\pm0.48}$ & $94.78_{\pm0.41}$ & $74.92_{\pm4.29}$ & $81.69_{\pm1.36}$ & $90.71_{\pm1.30}$ & $80.48_{\pm2.53}$ & $89.36_{\pm1.70}$ \\
        \DashedMidrule
        \rowcolor{Wheat1!60}
        \textsc{FreeDyG} + \methodname{} &  $\mathbf{98.43_{\pm0.02}}$ & $\mathbf{97.84_{\pm0.04}}$ & $\mathbf{84.02_{\pm0.31}}$ & $\mathbf{89.02_{\pm0.08}}$ & $\mathbf{94.44_{\pm0.25}}$ & $\mathbf{87.16_{\pm0.16}}$ & $\mathbf{92.82_{\pm0.15}}$ \\
        \bottomrule
    \end{tabular}
    \label{tab:inductive-ROC-Dyglib-2}
\end{table*}

\begin{table*}[t]
    \centering
    \small
    \setlength{\tabcolsep}{0.3pt}
    \renewcommand{\arraystretch}{1.0}
    \caption{Comparison of various neighborhood sampling methods on \emph{transductive} future edge prediction using 4 historical neighbors on different datasets from DyGLib. Performance is measured in ROC-AUC. The best performing method is marked in \textbf{bold}.}
    \begin{tabular}{l c c c c c c c}
        \toprule
         \multirow{2}*{Method $\downarrow$ / Dataset $\rightarrow$} & \textsc{Wikipedia}  & \textsc{Reddit} & \textsc{Mooc} & \textsc{LastFM} & \textsc{Social Evo.}  & \textsc{Enron}  & \textsc{UCI}  \\
         & \textsc{ROC-AUC} $\uparrow$ & \textsc{ROC-AUC} $\uparrow$ & \textsc{ROC-AUC} $\uparrow$ & \textsc{ROC-AUC} $\uparrow$ & \textsc{ROC-AUC} $\uparrow$ & \textsc{ROC-AUC} $\uparrow$ & \textsc{ROC-AUC} $\uparrow$\\
        \midrule
        \textsc{TGAT} + Trunc. & $93.11_{\pm0.41}$ & $94.36_{\pm0.24}$ & $81.28_{\pm0.14}$ & $62.99_{\pm1.04}$ & $91.12_{\pm0.16}$ & $67.21_{\pm0.56}$ & $75.99_{\pm1.10}$  \\
        \textsc{TGAT} + Uni. & $68.35_{\pm0.28}$ & $91.98_{\pm0.04}$ & $64.55_{\pm0.06}$ & $50.08_{\pm0.10}$ & $57.13_{\pm0.04}$ & $53.00_{\pm0.26}$ & $68.53_{\pm0.55}$ \\
        \textsc{TGAT} + NLB & $89.57_{\pm0.39}$ & $93.87_{\pm0.36}$ & $78.84_{\pm0.17}$ & $61.46_{\pm1.08}$ & $88.16_{\pm0.17}$ & $66.86_{\pm1.10}$ & $76.23_{\pm0.82}$ \\ 
                        \textsc{TGAT} + \textsc{TASER}  & $72.25_{\pm0.30}$ & $92.55_{\pm0.38}$ & $70.83_{\pm0.44}$ & $52.25_{\pm1.12}$ & $76.43_{\pm1.40}$ & $59.06_{\pm2.49}$ & $75.13_{\pm0.94}$ \\
        \DashedMidrule
        \rowcolor{Wheat1!60}
        \textsc{TGAT} + \methodname{} &  $\mathbf{93.96_{\pm0.54}}$ & $\mathbf{96.00_{\pm0.11}}$ & $\mathbf{81.52_{\pm0.32}}$ & $\mathbf{64.79_{\pm7.65}}$ & $\mathbf{94.10_{\pm0.19}}$ & $\mathbf{73.30_{\pm0.70}}$ & $\mathbf{77.04_{\pm1.65}}$ \\
        \midrule
        \textsc{TGN} + Trunc. & $98.42_{\pm0.08}$ & $98.58_{\pm0.03}$ & $91.77_{\pm0.68}$ & $79.48_{\pm2.16}$ & $94.66_{\pm0.39}$ & $88.93_{\pm1.50}$ & $92.86_{\pm0.41}$ \\
        \textsc{TGN} + Uni. & $96.73_{\pm0.26}$ & $98.61_{\pm0.04}$ & $86.95_{\pm0.98}$ & $68.86_{\pm2.36}$ & $77.91_{\pm7.29}$ & $88.40_{\pm1.18}$ & $90.00_{\pm1.54}$ \\
        \textsc{TGN} + NLB & $98.25_{\pm0.08}$ & $98.63_{\pm0.05}$ & $\mathbf{92.45_{\pm0.48}}$ & $76.16_{\pm3.51}$ & $93.73_{\pm0.62}$ & $89.15_{\pm0.33}$ & $92.45_{\pm0.52}$ \\

        \textsc{TGN} + \textsc{TASER}  & $98.18_{\pm0.10}$ & $98.62_{\pm0.01}$ & $92.16_{\pm0.73}$ & $75.98_{\pm1.96}$ & $94.90_{\pm0.59}$ & $88.38_{\pm0.72}$ & $92.46_{\pm0.45}$ \\
        \DashedMidrule
        \rowcolor{Wheat1!60}
        \textsc{TGN} + \methodname{}  & $\mathbf{98.59_{\pm0.03}}$ & $\mathbf{98.96_{\pm0.02}}$ & $92.26_{\pm0.79}$ & $\mathbf{86.41_{\pm1.09}}$ & $\mathbf{95.50_{\pm0.15}}$ & $\mathbf{90.26_{\pm1.31}}$ & $\mathbf{93.91_{\pm0.51}}$ \\
        \midrule
        \textsc{GraphMixer} + Trunc. & $95.98_{\pm0.05}$ & $95.82_{\pm0.20}$ & $82.61_{\pm0.18}$ & $72.58_{\pm0.16}$ & $92.50_{\pm0.08}$ & $83.90_{\pm0.09}$ & $\mathbf{90.66_{\pm0.83}}$ \\
        \textsc{GraphMixer} + Uni. & $80.96_{\pm0.33}$ & $93.36_{\pm0.06}$ & $71.54_{\pm0.10}$ & $62.45_{\pm0.27}$ & $58.59_{\pm0.42}$ & $59.22_{\pm1.49}$ & $72.13_{\pm2.42}$ \\
        \textsc{GraphMixer} + NLB & $95.22_{\pm0.07}$ & $96.21_{\pm0.02}$ & $80.98_{\pm0.07}$ & $72.55_{\pm0.14}$ & $90.96_{\pm0.05}$ & $84.34_{\pm0.34}$ & $89.29_{\pm0.77}$ \\
                     \textsc{GraphMixer} + \textsc{TASER}  & $94.83_{\pm0.83}$ & $95.90_{\pm0.11}$ & $81.44_{\pm2.59}$ & $72.64_{\pm0.22}$ & $91.49_{\pm1.14}$ & $82.79_{\pm0.27}$ & $88.29_{\pm0.69}$ \\

        \DashedMidrule
        \rowcolor{Wheat1!60}
        \textsc{GraphMixer} + \methodname{}  & $\mathbf{96.71_{\pm0.34}}$ & $\mathbf{97.05_{\pm0.04}}$ & $\mathbf{82.67_{\pm0.23}}$ & $\mathbf{84.50_{\pm0.53}}$ & $\mathbf{95.51_{\pm0.06}}$ & $\mathbf{85.50_{\pm0.79}}$ & $90.45_{\pm0.90}$ \\
        \midrule
        \textsc{DyGFormer} + Trunc. & $97.71_{\pm0.04}$ & $97.02_{\pm0.06}$ & $84.96_{\pm0.10}$ & $74.75_{\pm0.28}$ & $94.50_{\pm0.02}$ & $82.46_{\pm0.35}$ & $91.11_{\pm0.25}$ \\
        \textsc{DyGFormer} + Uni. & $97.66_{\pm0.07}$ & $96.97_{\pm0.03}$ & $84.91_{\pm0.16}$ & $74.94_{\pm0.20}$ & $94.51_{\pm0.03}$ & $82.60_{\pm0.38}$ & $90.56_{\pm0.11}$ \\
        \textsc{DyGFormer} + NLB & $97.81_{\pm0.04}$ & $97.07_{\pm0.06}$ & $82.81_{\pm0.23}$ & $74.72_{\pm0.13}$ & $93.77_{\pm0.04}$ & $84.49_{\pm0.30}$ & $89.97_{\pm0.25}$ \\

        \textsc{DyGFormer} + \textsc{TASER}  & $96.63_{\pm0.39}$ & $96.52_{\pm0.23}$ & $84.19_{\pm0.30}$ & $73.99_{\pm0.87}$ & $94.27_{\pm0.17}$ & $82.35_{\pm0.33}$ & $90.34_{\pm0.36}$ \\
        \DashedMidrule
        \rowcolor{Wheat1!60}
        \textsc{DyGFormer} + \methodname{}  & $\mathbf{98.53_{\pm0.01}}$ & $\mathbf{98.46_{\pm0.02}}$ & $\mathbf{85.68_{\pm0.22}}$ & $\mathbf{84.63_{\pm0.09}}$ & $\mathbf{95.68_{\pm0.01}}$ & $\mathbf{88.87_{\pm0.42}}$ & $\mathbf{92.48_{\pm0.18}}$ \\
        \midrule
        \textsc{FreeDyG} + Trunc. & $98.61_{\pm0.01}$ & $98.04_{\pm0.01}$ & $\mathbf{86.65_{\pm0.06}}$ & $79.93_{\pm0.05}$ & $95.19_{\pm0.01}$ & $90.58_{\pm0.10}$ & $94.30_{\pm0.20}$ \\
        \textsc{FreeDyG} + Uni. & $96.91_{\pm0.04}$ & $97.26_{\pm0.05}$ & $73.95_{\pm0.15}$ & $66.18_{\pm0.21}$ & $75.23_{\pm0.04}$ & \textsc{n/a} & $85.27_{\pm0.18}$ \\
        \textsc{FreeDyG} + NLB & $98.66_{\pm0.01}$ & $98.21_{\pm0.01}$ & \textsc{n/a} & $80.03_{\pm0.06}$ & $94.70_{\pm0.01}$ & $91.43_{\pm0.24}$ & $95.10_{\pm0.24}$ \\
           
        \textsc{FreeDyG} + \textsc{TASER}  & $98.37_{\pm0.27}$ & $98.00_{\pm0.09}$ & $83.08_{\pm2.81}$ & $80.14_{\pm0.67}$ & $94.52_{\pm0.29}$ & $90.66_{\pm0.47}$ & $92.87_{\pm0.56}$ \\

        \DashedMidrule
        \rowcolor{Wheat1!60}
        \textsc{FreeDyG} + \methodname{}  &  $\mathbf{99.04_{\pm0.02}}$ & $\mathbf{98.84_{\pm0.01}}$ & $86.31_{\pm0.22}$ & $\mathbf{87.19_{\pm0.16}}$ & $\mathbf{95.96_{\pm0.04}}$ & $\mathbf{92.50_{\pm0.20}}$ & $\mathbf{95.36_{\pm0.28}}$ \\
        \bottomrule
    \end{tabular}
    \label{tab:transductive-ROC-Dyglib-4}
\end{table*}

\begin{table*}[t]
    \centering
    \small
    \setlength{\tabcolsep}{0.3pt}
    \renewcommand{\arraystretch}{1.0}
    \caption{Comparison of various neighborhood sampling methods on \emph{inductive} future edge prediction using 4 historical neighbors on different datasets from DyGLib. Performance is measured in ROC-AUC. The best performing method is marked in \textbf{bold}.}
    \begin{tabular}{l c c c c c c c}
        \toprule
         \multirow{2}*{Method $\downarrow$ / Dataset $\rightarrow$} & \textsc{Wikipedia}  & \textsc{Reddit} & \textsc{Mooc} & \textsc{LastFM} & \textsc{Social Evo.}  & \textsc{Enron}  & \textsc{UCI}  \\
         & \textsc{ROC-AUC} $\uparrow$ & \textsc{ROC-AUC} $\uparrow$ & \textsc{ROC-AUC} $\uparrow$ & \textsc{ROC-AUC} $\uparrow$ & \textsc{ROC-AUC} $\uparrow$ & \textsc{ROC-AUC} $\uparrow$ & \textsc{ROC-AUC} $\uparrow$\\
        \midrule
        \textsc{TGAT} + Trunc. & $93.46_{\pm0.35}$ & $91.91_{\pm0.24}$ & $79.77_{\pm0.17}$ & $\mathbf{69.91_{\pm1.29}}$ & $89.55_{\pm0.19}$ & $64.30_{\pm1.04}$ & $76.88_{\pm1.04}$  \\
        \textsc{TGAT} + Uni. & $68.54_{\pm0.55}$ & $87.59_{\pm0.07}$ & $61.91_{\pm0.04}$ & $50.07_{\pm0.22}$ & $55.40_{\pm0.21}$ & $50.60_{\pm1.02}$ & $63.44_{\pm0.49}$ \\
        \textsc{TGAT} + NLB & $89.90_{\pm0.57}$ & $91.10_{\pm0.35}$ & $76.80_{\pm0.22}$ & $67.99_{\pm1.48}$ & $85.93_{\pm0.14}$ & $64.81_{\pm1.53}$ & $75.82_{\pm0.94}$ \\ 
                        \textsc{TGAT} + TASER & $71.86_{\pm0.25}$ & $88.57_{\pm0.41}$ & $67.26_{\pm0.49}$ & $53.90_{\pm2.12}$ & $73.23_{\pm1.39}$ & $56.27_{\pm1.96}$ & $67.75_{\pm0.80}$ \\
        \DashedMidrule
        \rowcolor{Wheat1!60}
        \textsc{TGAT} + \methodname{} &  $\mathbf{94.05_{\pm0.48}}$ & $\mathbf{94.15_{\pm0.16}}$ & $\mathbf{79.91_{\pm0.31}}$ & $69.17_{\pm9.39}$ & $\mathbf{92.36_{\pm0.61}}$ & $\mathbf{70.48_{\pm1.47}}$ & $\mathbf{77.76_{\pm1.36}}$\\
        \midrule
        \textsc{TGN} + Trunc. & $97.73_{\pm0.06}$ & $97.15_{\pm0.10}$ & $91.23_{\pm1.02}$ & $84.31_{\pm2.22}$ & $92.22_{\pm1.64}$ & $81.11_{\pm1.90}$ & $87.53_{\pm0.56}$\\
        \textsc{TGN} + Uni. & $95.57_{\pm0.20}$ & $97.13_{\pm0.15}$ & $84.35_{\pm1.15}$ & $77.93_{\pm2.56}$ & $66.49_{\pm1.97}$ & $79.37_{\pm3.15}$ & $80.40_{\pm3.17}$ \\
        \textsc{TGN} + NLB & $97.48_{\pm0.13}$ & $97.31_{\pm0.10}$ & $\mathbf{91.70_{\pm0.54}}$ & $82.18_{\pm2.94}$ & $91.10_{\pm0.99}$ & $80.38_{\pm1.36}$ & $86.89_{\pm0.87}$ \\

\textsc{TGN} + TASER & $97.37_{\pm0.14}$ & $97.27_{\pm0.12}$ & $91.58_{\pm0.61}$ & $80.40_{\pm2.09}$ & $93.16_{\pm1.13}$ & $79.51_{\pm0.86}$ & $87.03_{\pm0.18}$ \\
        \DashedMidrule
        \rowcolor{Wheat1!60}
        \textsc{TGN} + \methodname{} &  $\mathbf{97.94_{\pm0.04}}$ & $\mathbf{97.96_{\pm0.07}}$ & $91.16_{\pm1.10}$ & $\mathbf{88.24_{\pm1.39}}$ & $\mathbf{93.49_{\pm0.68}}$ & $\mathbf{83.04_{\pm2.52}}$ & $\mathbf{89.80_{\pm0.51}}$  \\
        \midrule
        \textsc{GraphMixer} + Trunc. & $95.43_{\pm0.04}$ & $93.36_{\pm0.16}$ & $81.33_{\pm0.19}$ & $79.47_{\pm0.18}$ & $91.24_{\pm0.12}$ & $76.45_{\pm0.43}$ & $\mathbf{89.41_{\pm0.50}}$ \\
        \textsc{GraphMixer} + Uni. & $81.10_{\pm0.28}$ & $88.67_{\pm0.05}$ & $71.92_{\pm0.26}$ & $71.45_{\pm0.24}$ & $49.76_{\pm0.27}$ & $50.38_{\pm1.14}$ & $69.34_{\pm2.36}$ \\
        \textsc{GraphMixer} + NLB & $94.34_{\pm0.14}$ & $93.58_{\pm0.07}$ & $79.15_{\pm0.17}$ & $79.48_{\pm0.19}$ & $89.15_{\pm0.13}$ & $76.25_{\pm0.87}$ & $87.99_{\pm0.48}$ \\

\textsc{GraphMixer} + TASER & $94.06_{\pm0.88}$ & $92.70_{\pm0.28}$ & $79.75_{\pm3.05}$ & $79.98_{\pm0.27}$ & $89.17_{\pm1.41}$ & $73.18_{\pm0.99}$ & $86.54_{\pm1.22}$ \\
        \DashedMidrule
        \rowcolor{Wheat1!60}
        \textsc{GraphMixer} + \methodname{} & $\mathbf{96.23_{\pm0.35}}$ & $\mathbf{95.36_{\pm0.03}}$ & $\mathbf{81.47_{\pm0.25}}$ & $\mathbf{87.64_{\pm0.53}}$ & $\mathbf{94.00_{\pm0.17}}$ & $\mathbf{78.63_{\pm1.15}}$ & $89.05_{\pm0.61}$ \\
        \midrule
        \textsc{DyGFormer} + Trunc. & $97.61_{\pm0.05}$ & $95.74_{\pm0.09}$ & $83.72_{\pm0.13}$ & $79.82_{\pm0.36}$ & $93.24_{\pm0.06}$ & $81.18_{\pm0.46}$ & $90.42_{\pm0.21}$ \\
        \textsc{DyGFormer} + Uni. & $97.56_{\pm0.06}$ & $95.71_{\pm0.04}$ & $83.71_{\pm0.18}$ & $80.08_{\pm0.25}$ & $93.24_{\pm0.06}$ & $81.42_{\pm0.70}$ & $90.01_{\pm0.13}$ \\
        \textsc{DyGFormer} + NLB & $97.62_{\pm0.02}$ & $95.79_{\pm0.10}$ & $81.27_{\pm0.25}$ & $79.86_{\pm0.21}$ & $92.39_{\pm0.07}$ & $82.97_{\pm0.47}$ & $89.30_{\pm0.25}$ \\

\textsc{DyGFormer} + TASER & $96.54_{\pm0.24}$ & $94.76_{\pm0.35}$ & $82.93_{\pm0.34}$ & $79.31_{\pm0.84}$ & $92.87_{\pm0.19}$ & $79.61_{\pm0.48}$ & $86.92_{\pm0.39}$ \\
        \DashedMidrule
        \rowcolor{Wheat1!60}
        \textsc{DyGFormer} + \methodname{} & $\mathbf{98.21_{\pm0.04}}$ & $\mathbf{97.83_{\pm0.02}}$ & $\mathbf{84.19_{\pm0.24}}$ & $\mathbf{87.67_{\pm0.08}}$ & $\mathbf{94.36_{\pm0.10}}$ & $\mathbf{86.13_{\pm0.84}}$ & $\mathbf{91.70_{\pm0.21}}$ \\
        \midrule
        \textsc{FreeDyG} + Trunc. & $98.23_{\pm0.02}$ & $97.05_{\pm0.02}$ & $\mathbf{85.54_{\pm0.11}}$ & $84.89_{\pm0.06}$ & $94.12_{\pm0.02}$ & $85.17_{\pm0.44}$ & $92.81_{\pm0.26}$ \\
        \textsc{FreeDyG} + Uni. & $96.70_{\pm0.03}$ & $95.53_{\pm0.07}$ & $74.64_{\pm0.17}$ & $73.94_{\pm0.10}$ & $69.12_{\pm0.15}$ & \textsc{n/a} & $80.65_{\pm0.25}$ \\
        \textsc{FreeDyG} + NLB & $98.23_{\pm0.03}$ & $97.25_{\pm0.04}$ & \textsc{n/a} & $85.07_{\pm0.06}$ & $93.46_{\pm0.04}$ & $86.16_{\pm0.24}$ & $92.81_{\pm0.26}$ \\

\textsc{FreeDyG} + TASER & $98.09_{\pm0.19}$ & $96.88_{\pm0.14}$ & $81.58_{\pm3.18}$ & $85.47_{\pm0.52}$ & $93.25_{\pm0.34}$ & $84.47_{\pm0.57}$ & $90.70_{\pm0.64}$ \\

        \DashedMidrule
        \rowcolor{Wheat1!60}
        \textsc{FreeDyG} + \methodname{} & $\mathbf{98.61_{\pm0.03}}$ & $\mathbf{98.28_{\pm0.01}}$ & $84.61_{\pm0.32}$ & $\mathbf{89.82_{\pm0.19}}$ & $\mathbf{94.36_{\pm0.33}}$ & $\mathbf{87.78_{\pm0.91}}$ & $\mathbf{93.49_{\pm0.18}}$ \\
        \bottomrule
    \end{tabular}
    \label{tab:inductive-ROC-Dyglib-4}
\end{table*}

\begin{table*}[t]
    \centering
    \small
    \setlength{\tabcolsep}{2.5pt}
    \renewcommand{\arraystretch}{1.0}
    \caption{Comparison of various neighborhood sampling methods on \emph{inductive} future edge prediction using 4 historical neighbors on different datasets from DyGLib. Performance is measured in AP. The best performing method is marked in \textbf{bold}.}
    \begin{tabular}{l c c c c c c c}
        \toprule
         \multirow{2}*{Method $\downarrow$ / Dataset $\rightarrow$} & \textsc{Wikipedia}  & \textsc{Reddit} & \textsc{Mooc} & \textsc{LastFM} & \textsc{Social Evo.}  & \textsc{Enron}  & \textsc{UCI}  \\
         & \textsc{AP} $\uparrow$ & \textsc{AP} $\uparrow$ & \textsc{AP} $\uparrow$ & \textsc{AP} $\uparrow$ & \textsc{AP} $\uparrow$ & \textsc{AP} $\uparrow$ & \textsc{AP} $\uparrow$\\
        \midrule
        \textsc{TGAT} + Trunc. & $94.42_{\pm0.34}$ & $92.45_{\pm0.17}$ & $78.81_{\pm0.21}$ & $72.33_{\pm1.27}$ & $87.35_{\pm0.25}$ & $68.41_{\pm0.72}$ & $79.99_{\pm1.00}$ \\
        \textsc{TGAT} + Uni. & $70.20_{\pm0.52}$ & $87.93_{\pm0.05}$ & $59.78_{\pm0.04}$ & $50.80_{\pm0.18}$ & $54.80_{\pm0.37}$ & $51.48_{\pm0.95}$ & $62.01_{\pm0.47}$ \\
        \textsc{TGAT} + NLB & $91.30_{\pm0.42}$ & $91.73_{\pm0.34}$ & $75.86_{\pm0.16}$ & $70.40_{\pm1.30}$ & $83.68_{\pm0.18}$ & $67.06_{\pm1.43}$ & $78.32_{\pm0.98}$ \\
        \textsc{TGAT} + TASER & $74.83_{\pm0.15}$ & $89.10_{\pm0.46}$ & $65.59_{\pm0.47}$ & $55.95_{\pm1.78}$ & $71.83_{\pm0.73}$ & $58.76_{\pm1.77}$ & $68.17_{\pm0.72}$ \\
        \DashedMidrule
        \rowcolor{Wheat1!60}
        \textsc{TGAT} + \methodname{} & $\mathbf{95.02_{\pm0.49}}$ & $\mathbf{94.76_{\pm0.13}}$ & $\mathbf{78.98_{\pm0.33}}$ & $\mathbf{72.72_{\pm0.83}}$ & $\mathbf{90.13_{\pm0.49}}$ & $\mathbf{74.88_{\pm1.15}}$ & $\mathbf{80.66_{\pm1.30}}$ \\
        \midrule
        \textsc{TGN} + Trunc. & $97.86_{\pm0.08}$ & $97.26_{\pm0.09}$ & $89.52_{\pm1.50}$ & $84.40_{\pm2.09}$ & $89.55_{\pm2.07}$ & $80.24_{\pm1.69}$ & $88.85_{\pm0.61}$ \\
        \textsc{TGN} + Uni. & $95.79_{\pm0.21}$ & $97.20_{\pm0.15}$ & $81.13_{\pm1.44}$ & $78.01_{\pm3.04}$ & $62.27_{\pm1.06}$ & $77.05_{\pm3.06}$ & $79.26_{\pm3.36}$ \\
        \textsc{TGN} + NLB & $97.62_{\pm0.10}$ & $97.40_{\pm0.09}$ & $\mathbf{89.90_{\pm0.46}}$ & $81.90_{\pm3.06}$ & $88.80_{\pm1.03}$ & $78.55_{\pm1.34}$ & $88.41_{\pm0.53}$ \\
        \textsc{TGN}+ TASER & $97.49_{\pm0.13}$ & $97.36_{\pm0.11}$ & $\mathbf{89.90_{\pm0.56}}$ & $80.14_{\pm2.84}$ & $91.05_{\pm1.52}$ & $77.32_{\pm1.42}$ & $88.11_{\pm0.31}$ \\
        \DashedMidrule
        \rowcolor{Wheat1!60}
        \textsc{TGN} + \methodname{} & $\mathbf{98.06_{\pm0.05}}$ & $\mathbf{98.12_{\pm0.06}}$ & $\mathbf{89.90_{\pm1.04}}$ & $\mathbf{88.87_{\pm1.04}}$ & $\mathbf{91.29_{\pm0.69}}$ & $\mathbf{82.67_{\pm2.38}}$ & $\mathbf{91.13_{\pm0.66}}$ \\
        \midrule
        \textsc{GraphMixer} + Trunc. & $95.92_{\pm0.06}$ & $93.74_{\pm0.18}$ & $79.91_{\pm0.22}$ & $81.15_{\pm0.28}$ & $88.70_{\pm0.15}$ & $75.68_{\pm0.72}$ & $\mathbf{91.29_{\pm0.37}}$ \\
        \textsc{GraphMixer} + Uni. & $83.13_{\pm0.19}$ & $88.91_{\pm0.06}$ & $68.88_{\pm0.26}$ & $74.70_{\pm0.19}$ & $50.65_{\pm0.39}$ & $53.23_{\pm0.97}$ & $70.29_{\pm2.51}$ \\
        \textsc{GraphMixer} + NLB & $94.93_{\pm0.14}$ & $93.96_{\pm0.07}$ & $77.81_{\pm0.20}$ & $81.22_{\pm0.11}$ & $86.60_{\pm0.14}$ & $75.39_{\pm0.60}$ & $90.09_{\pm0.30}$ \\
        \textsc{GraphMixer} + TASER & $94.50_{\pm0.88}$ & $93.14_{\pm0.23}$ & $78.48_{\pm3.30}$ & $81.45_{\pm0.23}$ & $87.07_{\pm1.42}$ & $72.36_{\pm0.49}$ & $89.07_{\pm0.97}$ \\
        \DashedMidrule
        \rowcolor{Wheat1!60}
        \textsc{GraphMixer} + \methodname{} & $\mathbf{96.70_{\pm0.35}}$ & $\mathbf{95.62_{\pm0.03}}$ & $\mathbf{80.20_{\pm0.24}}$ & $\mathbf{89.76_{\pm0.42}}$ & $\mathbf{91.06_{\pm0.27}}$ & $\mathbf{79.60_{\pm1.05}}$ & $91.04_{\pm0.46}$ \\
        \midrule
        \textsc{DyGFormer} + Trunc. & $97.94_{\pm0.04}$ & $96.53_{\pm0.08}$ & $84.21_{\pm0.09}$ & $83.50_{\pm0.20}$ & $91.05_{\pm0.11}$ & $83.81_{\pm0.73}$ & $93.05_{\pm0.14}$ \\
        \textsc{DyGFormer} + Uni. & $97.90_{\pm0.05}$ & $96.52_{\pm0.04}$ & $84.21_{\pm0.14}$ & $83.55_{\pm0.26}$ & $91.03_{\pm0.07}$ & $84.33_{\pm0.48}$ & $92.78_{\pm0.10}$ \\
        \textsc{DyGFormer} + NLB & $97.94_{\pm0.02}$ & $96.57_{\pm0.09}$ & $82.11_{\pm0.19}$ & $83.40_{\pm0.11}$ & $90.41_{\pm0.07}$ & $85.11_{\pm0.41}$ & $92.21_{\pm0.23}$ \\
        \textsc{DyGFormer} + TASER & $97.05_{\pm0.23}$ & $95.74_{\pm0.32}$ & $83.28_{\pm0.40}$ & $82.80_{\pm0.80}$ & $90.96_{\pm0.24}$ & $82.20_{\pm0.69}$ & $90.16_{\pm0.38}$ \\
        \DashedMidrule
        \rowcolor{Wheat1!60}
        \textsc{DyGFormer} + \methodname{} & $\mathbf{98.39_{\pm0.04}}$ & $\mathbf{98.18_{\pm0.02}}$ & $\mathbf{84.67_{\pm0.12}}$ & $\mathbf{89.66_{\pm0.08}}$ & $\mathbf{91.40_{\pm0.24}}$ & $\mathbf{86.99_{\pm0.54}}$ & $\mathbf{93.94_{\pm0.11}}$ \\
        \midrule
        \textsc{FreeDyG} + Trunc. & $98.44_{\pm0.03}$ & $97.48_{\pm0.01}$ & $\mathbf{85.73_{\pm0.08}}$ & $87.38_{\pm0.10}$ & $\mathbf{91.72_{\pm0.03}}$ & $84.64_{\pm0.32}$ & $94.46_{\pm0.13}$ \\
        \textsc{FreeDyG} + Uni. & $97.31_{\pm0.04}$ & $96.29_{\pm0.05}$ & $72.31_{\pm0.18}$ & $78.17_{\pm0.08}$ & $69.87_{\pm0.19}$ & n/a & $84.09_{\pm0.21}$ \\
        \textsc{FreeDyG} + NLB & $98.44_{\pm0.04}$ & $97.63_{\pm0.03}$ & n/a & $87.38_{\pm0.06}$ & $91.19_{\pm0.08}$ & $85.27_{\pm0.43}$ & $94.45_{\pm0.16}$ \\
        \textsc{FreeDyG} + TASER & $98.32_{\pm0.18}$ & $97.32_{\pm0.14}$ & $81.55_{\pm3.01}$ & $87.55_{\pm0.40}$ & $90.95_{\pm0.38}$ & $84.36_{\pm0.72}$ & $92.85_{\pm0.61}$ \\
        \DashedMidrule
        \rowcolor{Wheat1!60}
        \textsc{FreeDyG} + \methodname{} & $\mathbf{98.73_{\pm0.02}}$ & $\mathbf{98.50_{\pm0.02}}$ & $85.06_{\pm0.19}$ & $\mathbf{91.03_{\pm0.16}}$ & $90.77_{\pm0.95}$ & $\mathbf{87.38_{\pm0.55}}$ & $\mathbf{95.11_{\pm0.08}}$ \\
        \bottomrule
    \end{tabular}
    \label{tab:inductive-AP-Dyglib-4}
\end{table*}

We also compared \methodname{} with RepeatMixer~\cite{zou2024repeat}, a newly suggested TGNN with a dedicated neighborhood selection mechanism. We compared the models in the transductive setting and inductive setting and report the results in \Cref{tab:repeatmixer}.

\begin{table*}[t]
    \centering
    \small
    \setlength{\tabcolsep}{2.5pt}
    \renewcommand{\arraystretch}{1.0}
    \caption{Comparison of \methodname{} combined with a TGNN to RepeatMixer on the \emph{transductive} and \emph{inductive} tasks for future edge prediction using 2 historical neighbors on different datasets from DyGLib reported in AP. The  best performing method is marked in \textbf{bold}.}
    \begin{tabular}{l c c c c c c c}
        \toprule
         \multirow{2}*{Method $\downarrow$ / Dataset $\rightarrow$} & \textsc{Wikipedia}  & \textsc{Reddit} & \textsc{Mooc} & \textsc{LastFM} & \textsc{Social Evo.}  & \textsc{Enron}  & \textsc{UCI}  \\
         & \textsc{AP} $\uparrow$ & \textsc{AP} $\uparrow$ & \textsc{AP} $\uparrow$ & \textsc{AP} $\uparrow$ & \textsc{AP} $\uparrow$ & \textsc{AP} $\uparrow$ & \textsc{AP} $\uparrow$\\
        \midrule
        \textsc{RepeatMixer-transductive} & $98.90_{\pm0.02}$ & $98.76_{\pm0.01}$ & $84.28_{\pm0.18}$ & $\mathbf{91.10}_{\mathbf{\pm0.03}}$ & $92.92_{\pm0.05}$ & $\mathbf{91.58}_{\mathbf{\pm0.22}}$ & $95.78_{\pm0.11}$  \\
        \methodname\textsc{-transductive} & $\mathbf{98.94}_{\mathbf{\pm0.04}}$ & $\mathbf{99.06}_{\mathbf{\pm0.03}}$ & $\mathbf{91.19}_{\mathbf{\pm0.51}}$ & $90.54_{\pm0.62}$ & $\mathbf{93.62}_{\mathbf{\pm0.05}}$ & $91.31_{\pm0.01}$ & $\mathbf{95.86}_{\mathbf{\pm0.16}}$ \\
        \midrule
        \textsc{RepeatMixer-inductive} & $98.48_{\pm0.02}$ & $98.24_{\pm0.03}$ & $83.28_{\pm0.20}$ & $\mathbf{92.41}_{\mathbf{\pm0.04}}$ & $91.36_{\pm0.14}$ & $\mathbf{88.40}_{\mathbf{\pm0.70}}$ & $94.37_{\pm0.12}$ \\ 
        \methodname\textsc{-inductive} & $\mathbf{98.59}_{\mathbf{\pm0.03}}$ & $98.24_{\pm0.07}$ & $\mathbf{90.50}_{\mathbf{\pm0.61}}$ & $92.04_{\pm0.43}$ & $\mathbf{92.06}_{\mathbf{\pm0.48}}$ & $87.13_{\pm0.62}$ & $\mathbf{94.60}_{\mathbf{\pm0.13}}$ \\ 
        \bottomrule
    \end{tabular}
    \label{tab:repeatmixer}
\end{table*}

\begin{figure}[!h]
\centering
\includegraphics[width=\linewidth]{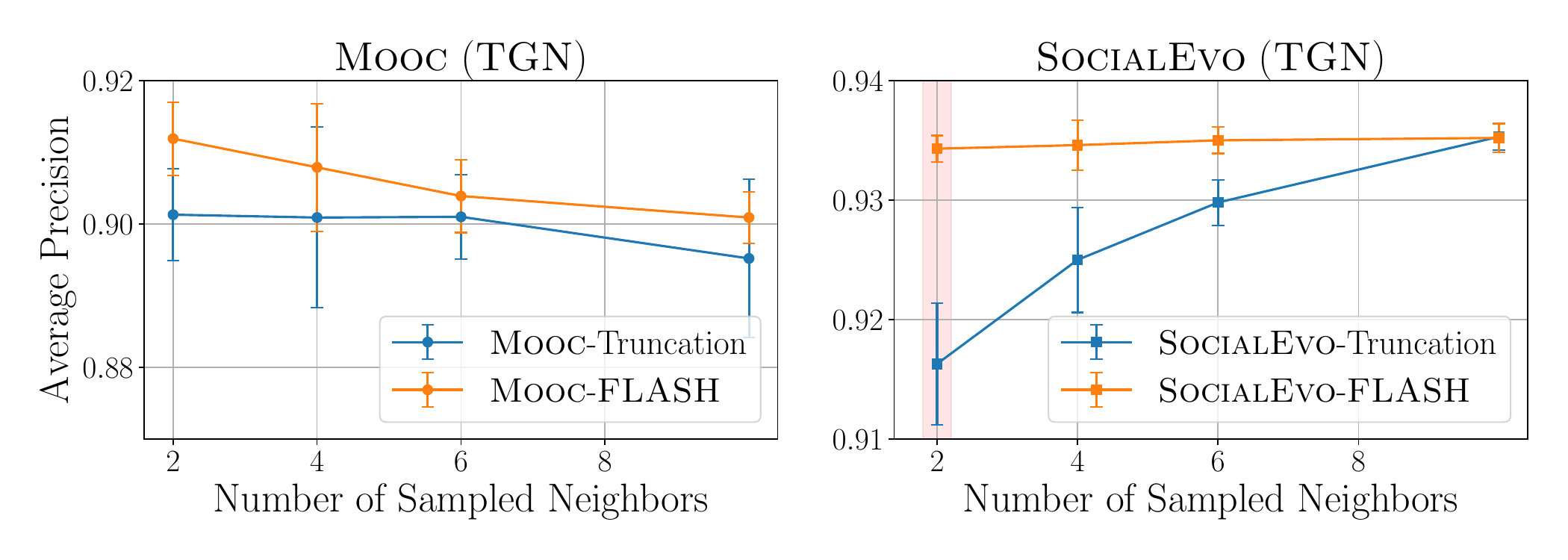}
\caption{Effect of neighbor count on \textsc{MOOC} (left) and \textsc{SocialEvo} (right). The bold, light‑red gap at 2 neighbors for SocialEvo shrinks as the sample size increases.}
\label{fig:ablation}
\end{figure}

\paragraph{Interpreting Predictive Relevance.} 
To justify the design of 
\methodname{}, we analyze its performance compared to truncation using TGN on \textsc{Mooc} and \textsc{Social Evo.}~\cite{yu2023towards}, as TGN is known to be sensitive to sampling strategy~\cite{rossi2020temporal}. In \Cref{fig:ablation}, we ablate on historical neighborhood sizes and conclude that \begin{enumerate*}[label=(\arabic*)]
\item when our method outperforms truncation, the most predictive nodes are not in the earliest part of the historical neighborhood, and 
\item when the performance is similar, the important nodes appear early.
\end{enumerate*}

%% file: ijcai26/appendix/appendixD.tex
\section{Additional experimental details}
\label{app:C}
We initialized $\gM$ with random numbers from a normal distribution with mean 0 and variance 1. In the inductive setting, the values of $\gM$ remain as initialized for the unseen nodes. We initialized each linear layer with random numbers from a uniform distribution within the range $[-\sqrt{d},\sqrt{d}]$ where $d$ is the dimension of the input to the linear layer.
We conducted a hyperparameter search for the dimension of $\gM$ over the set $\{10, 12, 16, 32\}$. In addition, recognizing that sampling from the entire historical neighborhood may introduce noise (as suggested by the results for \textsc{Mooc} in Figure 4), we fine-tuned the recent neighbors finder to sample either 10 or 32 historical neighbors. All other hyperparameters were consistent across models and matched those used in previous studies: a batch size of 200, the Adam optimizer, a learning rate of $10^{-4}$, and a binary cross-entropy loss function. When optimizing \ourmethod{}, if $\mathcal{S}_{v_i,v_j,t}$ overlaps with $\mathcal{S}_{v_i,t}^{\text{uni}}$ or  $\mathcal{S}_{v_j,v_i,t}$ overlaps with $\mathcal{S}_{v_j,t}^{\text{uni}}$, we resample the random neighborhoods to ensure an effective loss in \Cref{eqn:loss}.

In DyGLib, all models were trained for 100 epochs with early stopping after 20 epochs. For TGB, due to the lack of performance gains over many epochs on large-scale datasets (tgbl-review and tgbl-coin), we trained the models for only 3 epochs. For tgbl-wiki, each model was trained for 30 epochs.

To apply \methodname{} for a multi-hop neighborhood, one needs to first apply \methodname{} on the source or destination node, and then apply it again on each of  the sampled neighbors.

We used NVIDIA GeForce RTX 3090 and AMD Ryzen 9 7900X 12-Core Processor for our experiments.

%% file: ijcai26/appendix/appendixE.tex
\section{Theoretical Analysis}
\label{app:E}
For all proofs we assume that the TGNNs follows \Cref{eqn:H,eqn:z,eqn:merge}.
According to \Cref{th:truncation_fails}, there exists a dynamic graph such that any TGNN that utilize truncation sampling cannot learn this graph.
\begin{proof}
Let $\gG$ be a CTDG with $2k+1$ nodes, partitioned into three sets:
\[
A,\quad B,\quad \{v\},
\]
where $|A| = |B| = k$. We structure the events in $\gG$ over discrete timestamps $t \in \mathbb{N}$. At each timestamp $t$, $k$ new \emph{interactions} occur simultaneously: \begin{itemize}
    \item If $t \bmod 4 \in \{1,2\}$, then $v$ interacts with every node in $A$. 
    \item If $t \bmod 4 \in \{3,0\}$, then $v$ interacts with every node in $B$. 
\end{itemize}
A TGNN that utilize truncation sampling keeps only the last $k$ neighbors in each node’s historic neighborhood. Consequently, the sampled neighborhood for each node $a\in A$, $b\in B$ and $v$ we get 

\begin{table}[t]

\centering

\caption{Summary of sampled neighborhoods from $\gG$.}
\label{tab:trun_sum}
\setlength{\tabcolsep}{3pt}       
\renewcommand{\arraystretch}{1.1} 

\begin{tabular}{@{}l ccccc@{}}
\toprule
$t \bmod 4$ &
$S^{\mathrm{tru}}_{v,t}(k)$ &
$S^{\mathrm{tru}}_{a,t}(k)$ &
$S^{\mathrm{tru}}_{b,t}(k)$ &
positive & negative \\
\midrule
$1$ & $\{b\mid b\in B \}$ & $\{v\}$ & $\{v\}$ & $a$ & $b$ \\
$2$ & $\{a\mid a\in A \}$ & $\{v\}$ & $\{v\}$ & $a$ & $b$ \\
$3$ & $\{a\mid a\in A \}$ & $\{v\}$ & $\{v\}$ & $b$ & $a$ \\
$0$ & $\{b\mid b\in B \}$ & $\{v\}$ & $\{v\}$ & $b$ & $a$ \\
\bottomrule
\end{tabular}
\end{table}
the sampled neighborhoods as detailed in Table 13. When predicting future edges for times $t\bmod 4 = 2$ and  $t\bmod 4 = 3$ the inputs to the TGNN are the same with respect to the prediction of an edge $(v,a)$ where $a\in A$, but at $t\bmod 4 = 2$ the model should predict that the edge is positive and at $t\bmod 4 = 3$ the model should predict that the edge is negative. The same happens for $t\bmod 4 = 1$ and $t\bmod 4 = 0$ and  $(v,b)$ where $b\in B$. Hence, any TGNN that utilize truncation cannot overfit on this graph.

\end{proof}

According to Theorem 2, there exists a dynamic graph on which any TGNN that use $k$ uniform sampling cannot learn.
\begin{proof}
    Let $\gG$ be a CTDG with exactly 3 nodes: $a, b$ and $c$. Events in $\gG$ happen only for $t\in\mathbb{N}$. Whenever $t$ is odd, $a$ interacts with $b$, and whenever $t$ is even, $a$ interacts with $c$. We mark $t_{odd}$ as an odd time before the interaction occurred and $t_{even}$ as even time before the interaction occurred.
    $\mathcal{S}^{uni}_{b, t}(k) =\mathcal{S}^{uni}_{c, t}(k)=\{a\}$ for any $t$, since $b$ and $c$ only interacted with $a$. For every $t_{odd}$, $\frac{\#c}{\#b}=1$ where $\#u$ is the number of appearances of $u$ in the neighborhood of $a$. For every $t_{even}$, $\lim_{t_{even}\to\infty}\frac{\#c}{\#b}=1$. Hence, for every $k$ and a distribution distance $\delta$ there exists $t_k$ such that for every $t_k<t_{even}$ the distance between the distribution of $\mathcal{S}^{uni}_{a, t_{even}}(k)$ and $\mathcal{S}^{uni}_{a, t_{odd}}(k)$ is smaller than $\delta$. In addition, for every future positive edge at time $t$, the same edge is negative at $t+1$. For every positive future edge at $t_k<t$ there exists a negative edge at $t+1$ with exactly the same source, destination, sampled neighborhood of the destination node and $S^{uni}_{a,t}(k)$ and $S^{uni}_{b,t}(k)$ are sampled from approximately the same distribution. Therefore, the accuracy of a TGNN utilizing uniform sampling on this graph, is bounded.
\end{proof} 

To prove Theorem 3 we first prove the following lemmas:
\begin{lemma}
\label{lem:gen_trun}
    \methodname{} generalizes truncation, i.e. there exists a set of weights such that \methodname{} selects neighbors exactly as truncation.
\end{lemma}
\begin{lemma}
\label{lem:gen_uni}
    \methodname{} generalizes uniform sampling, i.e. there exists a set of weights such that \methodname{} sample neighbors exactly as uniform sampling.
\end{lemma}
\begin{lemma}
\label{lem:win_trun}
    There exists a simple TGNN and set of weights for \methodname{} that can learn the graph from Theorem 1.
\end{lemma}
\begin{lemma}
\label{lem:win_uni}
        There exists a simple TGNN and set of weights for \methodname{} that can learn the graph from Theorem 2.
\end{lemma}

We prove each of the lemmas:
\begin{proof}
\textbf{\Cref{lem:gen_trun}}.
    $r_u$ is an input to \methodname{}. Setting all the learnable weights that process other inputs to the zero and the linear projections that process $r_u$ to $I$, results in $r_u$ at the output of \methodname{}. We can set the final weights to $-I$, resulting in the output $-r_u$ for each input neighbor $u$. Selecting the top scored neighbors combined with a negative ranking loss is equivalent to truncation.
\end{proof}

\begin{proof}
    \textbf{\Cref{lem:gen_uni}}. We can set the final weights of \methodname{} to be zeros, resulting in outputting 0 score for each neighbor. 
    Since each neighbor has the same score, a uniform random selection is applied to select the neighbors.
    \end{proof}

\begin{proof}
    \textbf{\Cref{lem:win_trun}}. Given the historical neighbors of $v$ with ranks $1$ $(u_1)$, $2$ $(u_2)$ and $k+1$ $(u_{k+1})$, the following TGNN can achieve 100\% accuracy when predicting an edge $(v,u)$. When $u\in A$ in the dynamic graph from the proof of Theorem 1:
    \begin{align}
p 
\;=\;
\begin{cases}
        1; (u_1, u_2 \in A, u_{k+1} \in B )\vee( u_1, u_2, u_{k+1} \in B)\\
        0; otherwise 
\end{cases}
    \end{align}
 and where $u\in B$ 
     \begin{align}
     p 
\;=\;
\begin{cases}
        1; (u_1, u_2 \in B, u_{k+1} \in A )\vee (u_1, u_2, u_{k+1} \in A)\\
        0; otherwise 
\end{cases}
    \end{align}
    We now need to show that there is a set of weights such that \methodname{} gives the nodes with ranks 1, 2, and $k+1$ the top scores.
    We can set all the weights that process inputs to \methodname{}, other than $r_u$, to zeros, and set the linear projections until the final $\mathrm{MLP}$ layer as $I$, such that the final $\mathrm{MLP}$ layer is only function of $r_u$.
     The polynomial $-(x-1)^2(x-2)^2(x-(k+1))^2$ is continuous, hence according to \citep{hornik1989multilayer} the final $\mathrm{MLP}$ layer of \methodname{} can approximate it. This polynomial achieves maximum value at exactly three point, 1, 2 and $k+1$. Hence, \methodname{} that selects the top-3 scored neighbors can learn to select the neighbors with ranks $1,2$ and $k+1$ as desired by the simple TGNN that can achieve 100\% accuracy.
    \end{proof}

\begin{proof}
    \textbf{\Cref{lem:win_uni}}. From \Cref{lem:gen_trun}, \methodname{} can perform truncation. When setting the number of selected neighbors to $1$, \methodname{} chooses the most recent neighbor that interacted. We mark this neighbor as $u_1$.  
    The following TGNN can that achieve 100\% when predicting an edge $(a,u)$ where $u\in\{b,c\}$ in the CTDG from Theorem 2:
        \begin{align}
             p 
\;=\;
        \begin{cases}
        1; (u=c \wedge u_1=b) \vee (u=b \wedge u_1=c)\\
        0; otherwise 
        \end{cases}
    \end{align}
    \end{proof}

Combining the all the lemmas with Theorem 1 and Theorem 2 we receive that \methodname{} is \textit{more expressive} than truncation and uniform sampling.

%% file: ijcai26/appendix/appendixF.tex
\section{Time Complexity Analysis}
\label{app:F}
In addition to Table 4, we also provide in \Cref{tab:complexity} a time complexity analysis of different historical neighbors sampling strategies and \methodname{}.

\begin{table}[t]
\centering
\caption{Time complexity analysis. $n$ is the number of historical neighbors to select from and $k$ is the number of historical neighbors to select.}
\begin{tabular}{l ccc}
\toprule
Strategies & Time Complexity \\
\midrule
\textsc{Truncation} & $O(1)$ \\
\textsc{Uniform}    & $O(k)$ \\
\textsc{NLB} & $O(1)$ \\
\textsc{TASER} & $O(n+k\mathrm{log}(n))$ \\
\textsc{\methodname{} (ours)}&$O(n)$ \\
\bottomrule
\end{tabular}
\label{tab:complexity}
\end{table}

Since \methodname{} requires the top k neighbors and not randomly sampling based on scores, \methodname{} can be implemented using median of medians quickselect and, therefore, achieve time complexity of $O(n)$. In contrast, TASER requires random sampling which requires Fenwick tree to achieve optimal time complexity of $O(n+k\mathrm{log}(n))$. Both Truncation and NLB achieves $O(1)$ complexity due to maintaining the selected historical neighbors upon each new update to the graph. The maintenance operation of NLB is computationally intensive compared to truncation. Uniform sampling can be done in $O(K)$ using Fisher–Yates shuffle. Since the computation of the score of a neighbor by \methodname{} is independent by the computation of the other neighbors, \methodname{} can be easily parallelized to achieve better throughput.

%% file: ijcai26/appendix/AppendixG.tex
\section{FLASH vs TASER Analysis}
\label{app:G}
In this section, we discuss the main similarities and core differences between TASER and \ourmethod{}.

Both \ourmethod{} and TASER are adaptive neighbor sampling strategies that are sensitive to the evolution of the graph through time allowing the sampling mechanism adapt to the new graph structure, in contrast to the common heuristics of Truncation and Uniform. In addition, they both utilize a neighbor finder to first reduce the number of neighbor candidates and achieve a reasonable runtime that is independent of the graph size. However there are some major architectural aspects that differentiate between the two. First, the computation of gradients when optimizing TASER varies between one TGNN to another depends on the aggregation scheme employed by the TGNN. \ourmethod{}, however, is independent of the TGNN aggregator or any other variation to perform optimization, hence,  making it seamlessly integratable with any TGNN.
Second, TASER is link‑agnostic: it computes neighbor scores solely from the historical neighborhood and does not condition on the queried (potential) edge. Consequently, if a TGNN must sample neighborhoods for a positive and a negative edge simultaneously, TASER will assign identical neighbor scores to both queries. By contrast, \ourmethod{} is link‑adaptive, it incorporates the potential edge information in addition to each neighbor’s features, so it can assign different scores to different queried edges, even when evaluated at the same time. Furthermore, \ourmethod{} deterministically selects the top‑scoring neighbors, whereas TASER samples neighbors with probability proportional to their scores, which tends to be less robust. 

%% file: ijcai26.bib
@article{rossi2020temporal,
  title={Temporal graph networks for deep learning on dynamic graphs},
  author={Rossi, Emanuele and Chamberlain, Ben and Frasca, Fabrizio and Eynard, Davide and Monti, Federico and Bronstein, Michael},
  journal      = {CoRR},
  volume       = {abs/2006.10637},
  year         = {2020}
}

@article{xu2020inductive,
  title={Inductive representation learning on temporal graphs},
  author={Xu, Da and Ruan, Chuanwei and Korpeoglu, Evren and Kumar, Sushant and Achan, Kannan},
  journal={8th International Conference on Learning Representations, {ICLR} 2020,
  Addis Ababa, Ethiopia, April 26-30},
  year         = {2020},
}

@article{huang2024temporal,
  title={Temporal graph benchmark for machine learning on temporal graphs},
  author={Huang, Shenyang and Poursafaei, Farimah and Danovitch, Jacob and Fey, Matthias and Hu, Weihua and Rossi, Emanuele and Leskovec, Jure and Bronstein, Michael and Rabusseau, Guillaume and Rabbany, Reihaneh},
  journal={Advances in Neural Information Processing Systems},
  volume={36},
  year={2023},
}

@article{cong2023we,
  title={Do We Really Need Complicated Model Architectures For Temporal Networks?},
  author={Cong, Weilin and Zhang, Si and Kang, Jian and Yuan, Baichuan and Wu, Hao and Zhou, Xin and Tong, Hanghang and Mahdavi, Mehrdad},
  journal={The Eleventh International Conference on Learning Representations,
  {ICLR} 2023, Kigali, Rwanda, May 1-5},
  year={2023}
}

@inproceedings{kumar2019predicting,
  title={Predicting dynamic embedding trajectory in temporal interaction networks},
  author={Kumar, Srijan and Zhang, Xikun and Leskovec, Jure},
  booktitle={Proceedings of the 25th ACM SIGKDD International Conference on Knowledge Discovery \& Data Mining},
  pages={1269--1278},
  year={2019}
}

@article{yu2023towards,
  title={Towards better dynamic graph learning: New architecture and unified library},
  author={Yu, Le and Sun, Leilei and Du, Bowen and Lv, Weifeng},
  journal={Advances in Neural Information Processing Systems},
  volume={36},
  pages={67686--67700},
  year={2023}
}

@article{pennebaker2001linguistic,
  title={Linguistic inquiry and word count: LIWC 2001},
  author={Pennebaker, James W and Francis, Martha E and Booth, Roger J},
  journal={Mahway: Lawrence Erlbaum Associates},
  volume={71},
  number={2001},
  pages={2001},
  year={2001}
}

@article{shetty2004enron,
  title={The Enron email dataset database schema and brief statistical report},
  author={Shetty, Jitesh and Adibi, Jafar},
  journal={Information sciences institute technical report, University of Southern California},
  volume={4},
  number={1},
  pages={120--128},
  year={2004}
}

@article{madan2011sensing,
  title={Sensing the" health state" of a community},
  author={Madan, Anmol and Cebrian, Manuel and Moturu, Sai and Farrahi, Katayoun and others},
  journal={IEEE Pervasive Computing},
  volume={11},
  number={4},
  pages={36--45},
  year={2011},
  publisher={IEEE}
}

@article{panzarasa2009patterns,
  title={Patterns and dynamics of users' behavior and interaction: Network analysis of an online community},
  author={Panzarasa, Pietro and Opsahl, Tore and Carley, Kathleen M},
  journal={Journal of the American Society for Information Science and Technology},
  volume={60},
  number={5},
  pages={911--932},
  year={2009},
  publisher={Wiley Online Library}
}

@article{kazemi2019time2vec,
  title={Time2vec: Learning a vector representation of time},
  author={Kazemi, Seyed Mehran and Goel, Rishab and Eghbali, Sepehr and Ramanan, Janahan and Sahota, Jaspreet and Thakur, Sanjay and Wu, Stella and Smyth, Cathal and Poupart, Pascal and Brubaker, Marcus},
  journal      = {CoRR},
  volume       = {abs/1907.05321},
  year         = {2019},
}

@inproceedings{tian2023freedyg,
  title={FreeDyG: Frequency Enhanced Continuous-Time Dynamic Graph Model for Link Prediction},
  author={Tian, Yuxing and Qi, Yiyan and Guo, Fan},
  booktitle={The Twelfth International Conference on Learning Representations},
  year={2023}
}

@inproceedings{burges2005learning,
  title={Learning to rank using gradient descent},
  author={Burges, Chris and Shaked, Tal and Renshaw, Erin and Lazier, Ari and Deeds, Matt and Hamilton, Nicole and Hullender, Greg},
  booktitle={Proceedings of the 22nd international conference on Machine learning},
  pages={89--96},
  year={2005}
}

@article{tolstikhin2021mlp,
  title={Mlp-mixer: An all-mlp architecture for vision},
  author={Tolstikhin, Ilya O and Houlsby, Neil and Kolesnikov, Alexander and Beyer, Lucas and Zhai, Xiaohua and Unterthiner, Thomas and Yung, Jessica and Steiner, Andreas and Keysers, Daniel and Uszkoreit, Jakob and others},
  journal={Advances in neural information processing systems},
  volume={34},
  pages={24261--24272},
  year={2021}
}

@article{dingdygmamba,
  title={DyGMamba: Efficiently Modeling Long-Term Temporal Dependency on Continuous-Time Dynamic Graphs with State Space Models},
  author={Ding, Zifeng and Li, Yifeng and He, Yuan and Norelli, Antonio and Wu, Jingcheng and Tresp, Volker and Bronstein, Michael M and Ma, Yunpu},
  journal={Transactions on Machine Learning Research},
  year={2025}
}

@inproceedings{luoscalable,
  title={Scalable and Efficient Temporal Graph Representation Learning via Forward Recent Sampling},
  author={Luo, Yuhong and Li, Pan},
  booktitle={The Third Learning on Graphs Conference},
  year ={2024}
}

@article{hamilton2017inductive,
  title={Inductive representation learning on large graphs},
  author={Hamilton, Will and Ying, Zhitao and Leskovec, Jure},
  journal={Advances in neural information processing systems},
  volume={30},
  year={2017}
}

@inproceedings{ying2018graph,
  title={Graph convolutional neural networks for web-scale recommender systems},
  author={Ying, Rex and He, Ruining and Chen, Kaifeng and Eksombatchai, Pong and Hamilton, William L and Leskovec, Jure},
  booktitle={Proceedings of the 24th ACM SIGKDD international conference on knowledge discovery \& data mining},
  pages={974--983},
  year={2018}
}

@inproceedings{zeng2019graphsaint,
  title={GraphSAINT: Graph Sampling Based Inductive Learning Method},
  author={Zeng, Hanqing and Zhou, Hongkuan and Srivastava, Ajitesh and Kannan, Rajgopal and Prasanna, Viktor},
  booktitle={International Conference on Learning Representations},
year={2020}
}

@inproceedings{chen2018fastgcn,
  title={FastGCN: Fast learning with graph convolu-tional networks via importance sampling},
  author={Chen, Jie and Ma, Tengfei and Xiao, Cao},
  booktitle={International Conference on Learning Representations},
  year={2018},
  organization={International Conference on Learning Representations, ICLR}
}

@inproceedings{chiang2019cluster,
  title={Cluster-gcn: An efficient algorithm for training deep and large graph convolutional networks},
  author={Chiang, Wei-Lin and Liu, Xuanqing and Si, Si and Li, Yang and Bengio, Samy and Hsieh, Cho-Jui},
  booktitle={Proceedings of the 25th ACM SIGKDD international conference on knowledge discovery \& data mining},
  pages={257--266},
  year={2019}
}

@inproceedings{ni2019justifying,
  title={Justifying recommendations using distantly-labeled reviews and fine-grained aspects},
  author={Ni, Jianmo and Li, Jiacheng and McAuley, Julian},
  booktitle={Proceedings of the 2019 conference on empirical methods in natural language processing and the 9th international joint conference on natural language processing (EMNLP-IJCNLP)},
  pages={188--197},
  year={2019}
}

@article{shamsi2022chartalist,
  title={Chartalist: Labeled graph datasets for utxo and account-based blockchains},
  author={Shamsi, Kiarash and Victor, Friedhelm and Kantarcioglu, Murat and Gel, Yulia and Akcora, Cuneyt G},
  journal={Advances in Neural Information Processing Systems},
  volume={35},
  pages={34926--34939},
  year={2022}
}

@misc{wang2022reinforcementlearningenhancedweighted,
      title={Reinforcement Learning Enhanced Weighted Sampling for Accurate Subgraph Counting on Fully Dynamic Graph Streams}, 
      author={Kaixin Wang and Cheng Long and Da Yan and Jie Zhang and H. V. Jagadish},
      year={2022},
      eprint={2211.06793},
      archivePrefix={arXiv},
      primaryClass={cs.DB},
      url={https://arxiv.org/abs/2211.06793}, 
}

@misc{younesian2024grapeslearningsamplegraphs,
      title={GRAPES: Learning to Sample Graphs for Scalable Graph Neural Networks}, 
      author={Taraneh Younesian and Daniel Daza and Emile van Krieken and Thiviyan Thanapalasingam and Peter Bloem},
      year={2024},
      eprint={2310.03399},
      archivePrefix={arXiv},
      primaryClass={cs.LG},
      url={https://arxiv.org/abs/2310.03399}, 
}

@inproceedings{NIPS2017_3f5ee243,
 author = {Vaswani, Ashish and Shazeer, Noam and Parmar, Niki and Uszkoreit, Jakob and Jones, Llion and Gomez, Aidan N and Kaiser, \L ukasz and Polosukhin, Illia},
 booktitle = {Advances in Neural Information Processing Systems},
 editor = {I. Guyon and U. Von Luxburg and S. Bengio and H. Wallach and R. Fergus and S. Vishwanathan and R. Garnett},
 pages = {},
 publisher = {Curran Associates, Inc.},
 title = {Attention is All you Need},
 url = {https://proceedings.neurips.cc/paper_files/paper/2017/file/3f5ee243547dee91fbd053c1c4a845aa-Paper.pdf},
 volume = {30},
 year = {2017}
}

@inproceedings{zou2024repeat,
  title={Repeat-Aware Neighbor Sampling for Dynamic Graph Learning},
  author={Zou, Tao and Mao, Yuhao and Ye, Junchen and Du, Bowen},
  booktitle={Proceedings of the 30th ACM SIGKDD Conference on Knowledge Discovery and Data Mining},
  pages={4722--4733},
  year={2024}
}

@inproceedings{gravina2024long,
  title={Long Range Propagation on Continuous-Time Dynamic Graphs},
  author={Gravina, Alessio and Lovisotto, Giulio and Gallicchio, Claudio and Bacciu, Davide and Grohnfeldt, Claas},
  booktitle={International Conference on Machine Learning},
  pages={16206--16225},
  year={2024},
  organization={PMLR}
}

@article{zhou2022tgl,
  title={Tgl: A general framework for temporal gnn training on billion-scale graphs},
  author={Zhou, Hongkuan and Zheng, Da and Nisa, Israt and Ioannidis, Vasileios and Song, Xiang and Karypis, George},
  journal={arXiv preprint arXiv:2203.14883},
  year={2022}
}

@inproceedings{xu2024timesgn,
  title={TimeSGN: Scalable and Effective Temporal Graph Neural Network},
  author={Xu, Yuanyuan and Zhang, Wenjie and Zhang, Ying and Orlowska, Maria and Lin, Xuemin},
  booktitle={2024 IEEE 40th International Conference on Data Engineering (ICDE)},
  pages={3297--3310},
  year={2024},
  organization={IEEE}
}

@article{hornik1989multilayer,
  title={Multilayer feedforward networks are universal approximators},
  author={Hornik, Kurt and Stinchcombe, Maxwell and White, Halbert},
  journal={Neural networks},
  volume={2},
  number={5},
  pages={359--366},
  year={1989},
  publisher={Elsevier}
}

@inproceedings{deng2024taser,
  title={TASER: Temporal adaptive sampling for fast and accurate dynamic graph representation learning},
  author={Deng, Gangda and Zhou, Hongkuan and Zeng, Hanqing and Xia, Yinglong and Leung, Christopher and Li, Jianbo and Kannan, Rajgopal and Prasanna, Viktor},
  booktitle={2024 IEEE International Parallel and Distributed Processing Symposium (IPDPS)},
  pages={926--937},
  year={2024},
  organization={IEEE}
}
